\documentclass{article}

\usepackage{arxiv}
\usepackage{amsmath,amssymb}
\usepackage[utf8]{inputenc} % allow utf-8 input
\usepackage[T1]{fontenc}    % use 8-bit T1 fonts
\usepackage{hyperref}       % hyperlinks
\usepackage{url}            % simple URL typesetting
\usepackage{booktabs}       % professional-quality tables
\usepackage{amsfonts}       % blackboard math symbols
\usepackage{nicefrac}       % compact symbols for 1/2, etc.
\usepackage{microtype}      % microtypography
\usepackage{lipsum}
\usepackage{graphicx}
\usepackage{algorithmic}
\usepackage{graphicx}
\usepackage{textcomp}
\usepackage{pifont}
\usepackage{xspace}
\usepackage{float}
\usepackage{multirow}
\usepackage{subcaption}
\usepackage{tikz}
\usepackage[table]{xcolor}
\usepackage{colortbl}
\usepackage{booktabs}
\usepackage{makecell}
\usepackage{bbding}
\usepackage{float}
\usepackage{bm}
\usepackage[most]{tcolorbox}
\tcbuselibrary{listings, breakable, skins}
\newcommand{\one}{\ding{182}\xspace}
\newcommand{\two}{\ding{183}\xspace}
\newcommand{\three}{\ding{184}\xspace}
\newcommand{\four}{\ding{185}\xspace}

\newcommand{\fname}{\ensuremath{\mathcal{LMSAE }}}

\title{Lightweight Multi-Scale Anomaly Detection for Resource-Constrained Edge Devices}

\author{
 Raheen Junaid Wani \\
  School of Information Technology\\
  Indian Institute of Technology, Delhi\\
  New Delhi, India 110016 \\
    \And
 Smruti R. Sarangi \\
  Department of Computer Science\\
  Indian Institute of Technology, Delhi\\
  New Delhi, India 110016 \\
}

\begin{document}
\maketitle
\begin{abstract}
Time-series anomaly detection is increasingly important in IoT systems, sensor networks, and edge monitoring applications, where models must operate under strict constraints on memory, latency, and power consumption. While recent deep-learning approaches have improved detection accuracy, many remain computationally expensive and often fail to capture subtle anomalies due to limited multi-scale sensitivity. Autoencoders are widely used for anomaly detection because they reconstruct normal patterns well, leading to elevated reconstruction errors for anomalous inputs. Their simplicity and efficiency also make them suitable lightweight backbones for handling multi-scale inputs. To address these challenges, we propose a Lightweight MultiScale AutoEncoder (\fname{}) network for univariate time-series anomaly detection, designed to be compact and computationally efficient. \fname{} leverages the Discrete Wavelet Transform (DWT) to extract multi-scale features and employs a multi-scale loss function to improve sensitivity to subtle or hidden anomalies. Experiments on benchmark datasets demonstrate competitive or superior detection performance despite using significantly fewer parameters and a model size of less than 500 KB. \fname{} also achieves low-latency, low-power inference on the NVIDIA Jetson Nano, with $9\times$ reduction in inference latency and $2\times$ reduction in power consumption, making it ideal for edge deployment. 
\end{abstract}

% keywords can be removed
\keywords{Edge Computing \and Anomaly Detection \and Deep Learning \and Time Series \and Low-Latency Inference}

% Main text
\section{Introduction}
\label{sec:intro}
The widespread deployment of IoT devices, sensor networks, and edge-AI systems has led to an increase in real-time data
across domains such as healthcare, manufacturing, energy and smart infrastructure.  In most of these applications, the
data generated by sensors can be naturally represented as time series: sequences of observations indexed over time.
Detecting anomalies in such data is crucial for ensuring system safety, maintaining reliability and avoiding costly
failures. Sadly, Time Series Anomaly Detection (TSAD) is fundamentally challenging due to the complex nature of
time series signals, which often exhibit strong temporal dependencies, non-stationary behavior, missing values and
unpredictable changes.  Traditional anomaly detection systems often rely on centralized cloud infrastructure, where
sensor data is transmitted for processing and inference. This introduces latency, bandwidth overhead, and dependency on reliable connectivity.
In contrast, edge computing enables data to be processed closer
to the source, reducing latency and reliance on cloud connectivity. Edge devices typically use machine learning (ML) at the network edge to gather sensor data from large connected devices, analyze the data they receive, and identify trends and various operational situations for real-time decision making (\cite{edge}). 

Traditional TSAD techniques
(\cite{traditional-methods}) based on statistical methods (\cite{spot}), distance/density methods (\cite{knn}) or clustering
methods find it difficult to identify the temporal patterns associated with time series. Deep learning techniques, on
the other hand, provide the capacity to naturally learn expressive, feature-rich representations. Specifically,
unsupervised approaches work well in real-world TSAD scenarios since labeled anomaly data is scarce and costly
to obtain. Reconstruction-based models (~\cite{lstm-encdec},~\cite{tsvae}, \cite{cvae}, ~\cite{tadgan}), have gained prominence for their ability to learn to reconstruct normal patterns and detect anomalies via reconstruction error. More recent works like
FCVAE (\cite{fcvae}) show that frequency representations can significantly improve anomaly detection for VAEs. However, these methods struggle to capture \emph{subtle anomalies}. 
 
\begin{figure*}
    \centering
    \includegraphics[width=1\linewidth]{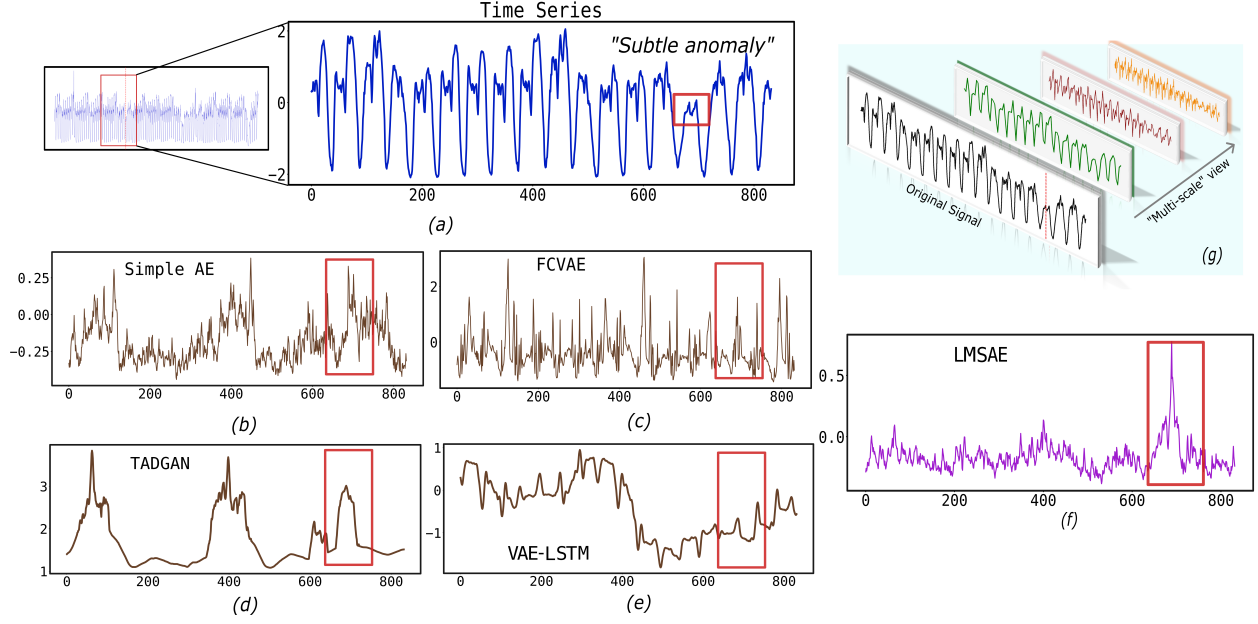}
    \caption{Time series anomaly detection comparison across different models. \emph{(a)} The input time series with a subtle anomaly highlighted with the x-axis representing the time index.
\emph{(b–f)} Anomaly scores from various models. Only \fname{} \emph{(f)} clearly detects the subtle anomaly whereas other models either miss it or exhibit noisy patterns. (\emph{g}) gives a visual idea of the multi-scale view.}
    \label{fig:anomalies}
\end{figure*}

Consider Figure \ref{fig:anomalies}(a) which shows a magnified portion of a time series dataset from an anomaly detection benchmark (\emph{NAB-NYC Taxi} (\cite{b2})). It represents taxi passenger data over time. Most of the signal exhibits a repeating daily pattern: regular rises and falls reflecting typical passenger volumes. A \emph{subtle anomaly} can be seen as a short, localized deviation that does not break this broad pattern drastically but still departs from the expected structure in a non-obvious way. For instance, the anomaly highlighted in red box does not show a very major spike or dip, yet it subtly disrupts the rhythm. Visually, it blends in with the rest of the series unless closely examined, and can be easily missed by models focused only on large scale changes. 
Figures \ref{fig:anomalies}(b) to (f) show how some of the reconstruction-based TSAD models, including ours (referred to as \fname{}), respond to this segment of the time series.
It can be observed that the anomaly score produced by these models (Figure \ref{fig:anomalies}(b-e)) for a window containing this anomaly remains similar to the anomaly scores for the normal points. They treat the anomaly just like any other normal variation because it does not stand out in terms of size or shape. From their perspective, nothing unusual is happening. However, Figure \ref{fig:anomalies}(f) shows the output from our approach, which clearly marks this region with a noticeable peak in the anomaly score. Even though the anomaly is not visually dramatic, our method detects that the rhythm of the signal has changed in a meaningful way. This kind of sensitivity is crucial, especially in real-world settings where early or weak anomalies can be the first sign of a developing issue. For example, a small yet unusual dip in passenger volume could reflect a minor system disruption, a localized event, or early signs of a broader pattern shift. If such anomalies are missed, they might accumulate or go unnoticed until they become more serious. Therefore, being able to reliably detect subtle deviations without relying on large-scale changes is essential not just for accuracy but also for timely and informed decision-making, especially in environments where quick responses matter.\\
Reconstruction-based methods (based on either AEs (Fig \ref{fig:anomalies}(b)), VAEs (Fig \ref{fig:anomalies}(c,e)) or GANs (Fig \ref{fig:anomalies}(d)), whether operating in the time or frequency domain, typically process the entire signal as a single stream. During training, they learn to reconstruct the overall shape of the time series by minimizing the average reconstruction error across all timesteps, due to which they tend to learn the large-scale patterns and effectively generalize the series. Since these models learn to generalize the signal’s dominant behavior, small or localized deviations often get smoothed out or absorbed into the reconstruction. This limitation stems from a lack of multi-scale sensitivity (refer to Fig \ref{fig:anomalies}(g) for an intuitive visual idea). Anomalies that are small in amplitude or confined to a narrow temporal or frequency band remain undetected unless they cause a significant disruption in the overall signal. Even when frequency cues are used, unless the anomaly leaves a strong enough footprint, it tends to be overlooked. Without an explicit way to decompose and isolate patterns at different resolutions, these models miss out on the nuances.

Recent advances in TSAD have steadily improved detection accuracy through increasingly complex model designs. However,
this evolution has also led to a sharp rise in model size and parameter count, shifting the field toward architectures
that are difficult to deploy on edge devices. For instance, the STM32WB55 MCU offers only 1 MB of flash memory and 256 KB of SRAM, which places tight
restrictions on model size. In this scenario, TSAD models like TADGAN (\cite{tadgan}) and
VAE-LSTM (\cite{tsvae}), with model sizes 1.1 MB and 2.8 MB respectively, are simply too large to be deployed on such devices. Even when using more capable edge-AI platforms like the NVIDIA Jetson Nano (which offers 2 GB or 4 GB RAM), there are high inference latency and energy consumption costs associated due to limited GPU compute capacity and power constraints ($5$W-$10$W). Models with new architectures, such as attention modules, are even larger, with FCVAE (\cite{fcvae}), having a model size of 13 MB. 
 Since existing models are built for powerful systems, they require significant architectural changes for edge deployment. In contrast, our model is designed from the ground up for both server-class performance and edge efficiency, enabling seamless deployment across platforms without the need for separate versions or post-hoc optimizations. 
 
Therefore, there are two main challenges that have persisted until now: \one The need to detect subtle anomalies that may be critical in real-time applications but are often missed by existing methods. \two Need to create an architecture that is not too bulky in terms of parameter count and model size, and that can perform well on powerful systems while also being deployable on resource-constrained edge devices. To address these, we propose a novel and resource-efficient model architecture called \textbf{\fname{}} (Lightweight MultiScale AutoEncoder). For the first challenge, the model is  designed to enhance sensitivity to \emph{subtle anomalies} by including a multi-resolution view where Discrete Wavelet Transforms are a part of the picture. For the second challenge, we use a lightweight autoencoder backbone to process different scales appropriately resulting in a robust anomaly detection model. In summary, our main contributions are as follows:

\begin{itemize}
    \item We introduce \fname{}, a compact TSAD architecture that augments a basic autoencoder with multi-scale wavelet decomposition, enhancing anomaly detection sensitivity, especially for subtle 
    deviations that disrupt the rhythm.
    \item We formulate a multi-scale loss function that enhances sensitivity to subtle anomalies. Specifically, an anomaly that affects a single  frequency level will result in a large error at that level, even if its impact on the overall signal is minor. This decomposition allows the system to act as if it has multiple "alarms", one for each scale, where a significant deviation in any level signals a potential anomaly.
    \item We significantly reduce the model size and the parameter count as compared to existing baselines: \fname{} uses only $97K$ parameters and a model size of $425 KB$, in contrast to TADGAN ($250K/1.1MB$), VAE-LSTM ($700K/2.8MB$), and FCVAE  \\
    ($1400K/13MB$).
    \item We extensively evaluate our model on standard anomaly detection benchmarks (NAB and Yahoo) and show
    competitive or superior performance despite a significantly smaller model size. 
    \item We demonstrate the feasibility of deployment on edge devices, such as the NVIDIA Jetson Nano. 
\end{itemize}

The paper is organized as follows. Section~\ref{sec:background} provides the background of Wavelet
Transform and Autoencoders. Sections~\ref{sec:problem} and \ref{sec:methodology} discusses the problem and the design method. Section~\ref{sec:experiments} provides the
details of the experimental setup. Sections~\ref{sec:results} and \ref{sec:interpretation} presents the experimental results and model analysis, respectively. We present the related work in 
Section~\ref{sec:relatedwork} and finally conclude in Section~\ref{sec:conclusion}.

\section{Background}
\label{sec:background}
\subsection{Wavelet Transforms}
\label{sec:wt}
The wavelet transform uses {\em wavelets} as basis functions, where a wavelet is defined as an oscillatory function with
a compact support (defined only for a small window of time).  Consider a mother wavelet satisfying the two conditions:
$\int_{-\infty}^{\infty} \psi(t)dt=0$ (area under the curve is zero) and $\int_{-\infty}^{\infty} |\psi(t)|^2dt < \infty
$ (finite energy). This can be scaled and shifted to generate different {\em child wavelets}: $\frac{1}{\sqrt{s}}
\psi(\frac{t-k}{s})$ where $s$ is the scaling parameter and $k$ is the time-shifting parameter. $s$
corresponds to the frequency information ($\frac{1}{f}$), which dilates or compresses the signal, while $k$ shifts
the wavelet along the signal (\cite{dwt-psi}, \cite{dwt-eq}).

The wavelet is shifted across time, filling the rows of the time-scale plane, by changing the translation $k$ while maintaining a constant scale $s$. Conversely, filling the columns of the time-scale plane with a fixed $k$ while varying the scale $s$ modifies the frequency resolution. Wavelet coefficients are then obtained by processing the signal to be studied using the child wavelets. Wavelet transforms can be either {\em continuous} (CWT) or {\em discrete} (DWT).  While DWTs employ a certain subset of scale and translation values, CWTs function across all potential scales and translations.

The Continuous Wavelet Transform (CWT) of an input signal $x(t)$ is defined as:
\begin{equation}
    W_{\psi}(s,k) = \frac{1}{\sqrt{s}} \int_{-\infty}^{\infty} x(t)  \psi^{*} \left( \frac{t-k}{s} \right) dt
\end{equation}
where, $\psi^*$ is the complex conjugate. The components in $W_{\psi}(s,k)$ are referred to as wavelet coefficients, and each one is linked to a frequency scale and a time domain point (\cite{wavelet-theory}). 

The Discrete Wavelet Transform (DWT)  discretizes these parameters using $(s=2^j, k=2^jn)$:

\begin{equation}
    W_{\psi}[j,n] = \frac{1}{\sqrt{2^j}} \sum_{m=-\infty}^{\infty} x[t_m]  \psi \left(2^{-j}t_m-n \right) dt
\end{equation}

Here, the integers $j,n$ control the dilation and translation respectively.  However, in practical implementations, the
true Discrete Wavelet Transform (DWT) does not rely on integration but rather uses filter banks for signal
decomposition. The filters can be interpreted as the wavelet functions at different scales as explained in the next
section.

In this way, a wavelet transform breaks a signal into different frequency components while preserving temporal
information. Unlike the Fourier transform, which gives us information in the frequency domain, the wavelet transform
also tells us {\em when} in time a signal with a given frequency occurs through the time-shifting property. The Fourier
transform offers a global perspective by converting the entire signal into its frequency spectrum, meaning that
localized frequency variations become part of a broad, overall representation. To account for this limitation, a window
based Fourier Transform or the more generalized Short-Time Fourier Transform (STFT) was introduced. The STFT applies a
window function of fixed length to the signal, computing the Fourier transform within each window. By shifting this
window across the signal, the STFT also provides a time-frequency representation. However, this approach introduces two
drawbacks: Firstly, the fixed window size may not be optimal for capturing both short-duration and long-duration signal
features. If a signal contains feature variations that are either much smaller or much larger than the window, important
details can be lost or misrepresented.  Secondly, the time resolution is same for high and low frequencies.  Wavelet
Transform overcomes these limitations by using the window that is scaled in both time and frequency as we can stretch or
compress the wavelet to analyze different frequency features with varying resolution. This adaptability makes wavelets
particularly useful for detecting changes, anomalies, and patterns across different time scales.
\subsection{Discrete Wavelet Transform} 
\label{sec:dwt}
The Discrete Wavelet Transform (DWT) (\cite{b1}) decomposes a discrete time series into \emph{approximate} and
\emph{detailed} coefficients using a series of high-pass and low-pass filters, followed by downsampling.  This process
is recursively applied to the approximate coefficients, generating a multiscale representation, where each level
corresponds to a different resolution of the original signal. High-pass and low-pass filters correspond to a wavelet
function $\psi(t) = \frac{1}{\sqrt{s}} \psi (\frac{t-k}{s})$ and a scaling function $\phi(t) = \frac{1}{\sqrt{s}} \phi
(\frac{t-k}{s})$, respectively, which collectively form an orthogonal basis (\cite{wave-rora}).  The low-pass and
high-pass filter coefficients depend on the choice of the basis function. The {\em Haar Wavelet} (\cite{haar}) and the
{\em Daubechies Wavelet} (\cite{daubechies}) are the commonly used wavelets. The Haar Wavelet is the simplest and is defined using: \( \phi(t) = \begin{cases} 
1, & 0 \leq t < 1, \\ 
0, & \text{otherwise} 
\end{cases} \) where $\phi(t)$ is the scaling function for low-pass filter and \( \psi(t) = \begin{cases} 
1, & 0 \leq t < \frac{1}{2}, \\ 
-1, & \frac{1}{2} \leq t < 1, \\ 
0, & \text{otherwise} 
\end{cases} \) where $\psi(t)$ is the wavelet function for the high-pass filter.

For a single-level decomposition, the input signal {\em x[n]} is passed through a high-pass and a low-pass filter,
represented by {\em g[n]} and {\em h[n]} respectively, followed by a downsampling operation. Filtering is performed
using convolution as:
%FIXME: explain this. What is the downarrow?
\begin{align}
\label{eq:low}
   y_{l}[n] &= (x \ast h) \downarrow 2  \nonumber \\
   &= \left( \sum_{k=-\infty}^{\infty} x[k] h[n-k] \right) \downarrow 2 \nonumber \\
    &= \sum_{k=-\infty}^{\infty} x[k] h[2n-k]
\end{align}

\begin{align}
\label{eq:high}
   y_{h}[n] &= (x \ast g) \downarrow 2 \nonumber\\
   &= \left( \sum_{k=-\infty}^{\infty} x[k] g[n-k] \right) \downarrow 2 \nonumber \\
    &= \sum_{k=-\infty}^{\infty} x[k] g[2n-k]
\end{align}

where ($\ast$) represents a convolution operation and ($\downarrow2$) represents downsampling by a factor of $2$:
$y[n]=x[2n]$. In other words, every sample index which is a multiple of $2$ is retained and others are discarded.
Similarly, for the $j^{th}$ level, the signal is split into a new set of approximation and detail coefficients using
approximation coefficient from the previous level. 

\begin{equation}
   y_{l}^{j+1}[n] = (y_{l}^j \ast h) \downarrow 2 
\end{equation}
\begin{equation}
   y_{h}^{j+1}[n] = (y_{l}^j \ast g) \downarrow 2 
\end{equation}
The frequency resolution increases as the number of levels increases, while the time resolution decreases due to downsampling. 

{\em Inverse DWT: } The original signal can also be reconstructed from the approximate and detailed coefficients at
every level.  The approximation and detail coefficients $(y_l, y_h)$ are first upsampled by a factor of 2, then passed
through the synthesis high-pass and low-pass filters denoted as {\em g'[n]} and {\em h'[n]} respectively. The synthesis
filters are often derived from the filters used in forward DWT, but their exact form depends on the wavelet family being
used. In some wavelet transforms, they may be identical to the orginal filters (e.g., Haar wavelet), while in others,
they may be modified versions, often involving time-reversal or scaling. A single-level reconstruction, given we have
$(y_l, y_h)$ from a single-level forward DWT using Equations  (\ref{eq:low}) and (\ref{eq:high}) can be written as:

\begin{align} 
x[n] &= (y_l[n] \uparrow 2) \ast h' + (y_h[n] \uparrow 2) \ast g' 
\end{align}

where $(\uparrow 2)$ represents an upsampling operation (insertion of zeros between samples), and is defined as:
\begin{equation}
\label{eq:upsample}
y[n] = \begin{cases} x[n/2], & n \text{ even} \\ 
0, & n \text{ odd} 
\end{cases} 
\end{equation}

Expanding the convolutions, the reconstruction equation becomes:

\begin{align} 
\label{eq:idwt}
x[n] &= \sum_{k=-\infty}^{\infty} y_l[k] h'[n - 2k] + \sum_{k=-\infty}^{\infty} y_h[k] g'[n - 2k] 
\end{align}
Similarly, for multi-level reconstruction, at the $jth$ level, the process iterates as:
\begin{equation} 
y_l^j[n] = (y_l^{j+1} \uparrow 2) \ast h' + (y_h^{j+1} \uparrow 2) \ast g' 
\end{equation} until we reach the first level where we get $x[n]$. 
 
\subsection{Autoencoders for Anomaly Detection}
Finding data points or patterns that substantially differ from the majority of
samples is known as time series anomaly detection. These irregularities may be
signs of fraud, cybersecurity risks, system malfunctions, or other odd
events. These anomalies could be \emph{point anomalies}, in which one
observation deviates significantly from the expected value; \emph{contextual
anomalies}, in which a value may be normal in one context but abnormal in
another; or \emph{collective anomalies}, in which a collection of values
exhibits an unexpected pattern. 

A neural network that can be used for such unsupervised learning tasks is an {\em Autoencoder}. An autoencoder consists
of two main components: an {\em encoder (E)} that maps the input data in to a lower-dimensional latent representation
and a {\em decoder (D)}, which reconstructs the original input from the latent representation.  Given an input sequence
\( x^{(i)} \in \mathbb{R}^N\), where \( N \) is the sequence length, the encoder compresses it into a latent
representation \( z^{(i)} \) of lower dimensionality \( d \), where \( d \ll N \):

\begin{equation}
    z^{(i)} = E_\theta(x^{(i)}), z^{(i)} \in \mathbb{R}^d, d \ll N
    \label{eq:enc}
\end{equation}

The decoder reconstructs the input from this latent space representation:

\begin{equation}
    \hat{x}^{(i)} = D_\theta(z^{(i)})
    \label{eq:dec}
\end{equation}

$E_{\theta}$ and $D_\theta$ represents the {\em Encoder} and {\em Decoder}
networks respectively, parameterized by $\theta$ and \( \hat{x}^{(i)} \) is the reconstructed
version of the input. The reconstruction error measures the difference between
the original input ($x$) and its reconstruction ($\hat{x}$). A common metric
used for this is the Mean Squared Error (MSE), computed as:

\begin{equation}
    \mathcal{L}(x^{(i)},\hat{x}^{(i)}) = \frac{1}{N}\sum_{j=1}^N(x_{i+j-1} - \hat{x}_{i+j-1})^2
    \label{eq:error}
\end{equation}
where \( \mathcal{L}(x^{(i)},\hat{x}^{(i)}) \) represents the reconstruction loss for the input \( x^{(i)} \). 

Because autoencoders encode and decode the input data, they can learn an
internal representation of normal patterns in a dataset. When encountering new
data that significantly deviates from these learned patterns, the reconstruction
process may fail, leading to a noticeable increase in the reconstruction error. This
property makes autoencoders particularly useful for anomaly detection in time
series datasets. The objective of training an {\em Autoencoder} is to minimize the
reconstruction loss, ensuring that the reconstructed output \( \hat{x}^{(i)} \)
is as close as possible to the original input \( x^{(i)} \). After the training
phase, the autoencoder can be used to detect anomalies in new time series data
by analyzing its reconstruction error. If an input sample follows the learned
normal patterns, the autoencoder can accurately reconstruct it, leading to a low
reconstruction error. If an input sample significantly deviates from normal
patterns such as an unexpected spike, missing values, or an unusual trend the
autoencoder struggles to reconstruct it accurately, resulting in a high
reconstruction error. A threshold can be set on the reconstruction error to
classify time series data points as normal or anomalous.

\section{Problem Formulation}
\label{sec:problem}
We formulate the task of UTSAD via reconstruction method as follows: consider a univariate time series sequence
$\mathbf{x}=\{x_1, x_2...x_N\}$ where $x_i$ represents the observed value at time $t$. Since time series data exhibit
temporal dependencies, we segment the time series into overlapping sliding windows of length {\em W} such that each
window is defined as $\mathbf{{x}^{(i)}}=\{x_i,x_{i+1},...,x_{i+W-1} \}$, $i=1,2,...N-W+1$. Each window
$\mathbf{x^{(i)}}$ is passed through an autoencoder model $f_\theta$ where $\theta$ represents the learnable parameters
of the model according to Equations \eqref{eq:enc} and \eqref{eq:dec}. Anomalies are detected by comparing the
reconstruction error given by Equation~\eqref{eq:error} to a predefined threshold $\tau$: $\mathbf{x}^{(i)}$ is an
anomaly if $\mathcal{L}(\mathbf{x^{(i)}},\mathbf{\hat{x}^{(i)}}) > \tau$ (\cite{anomaly-review}).

\section{Design Methodology} 
\label{sec:methodology}
\begin{figure*}
    \centering
    \includegraphics[width=1\linewidth]{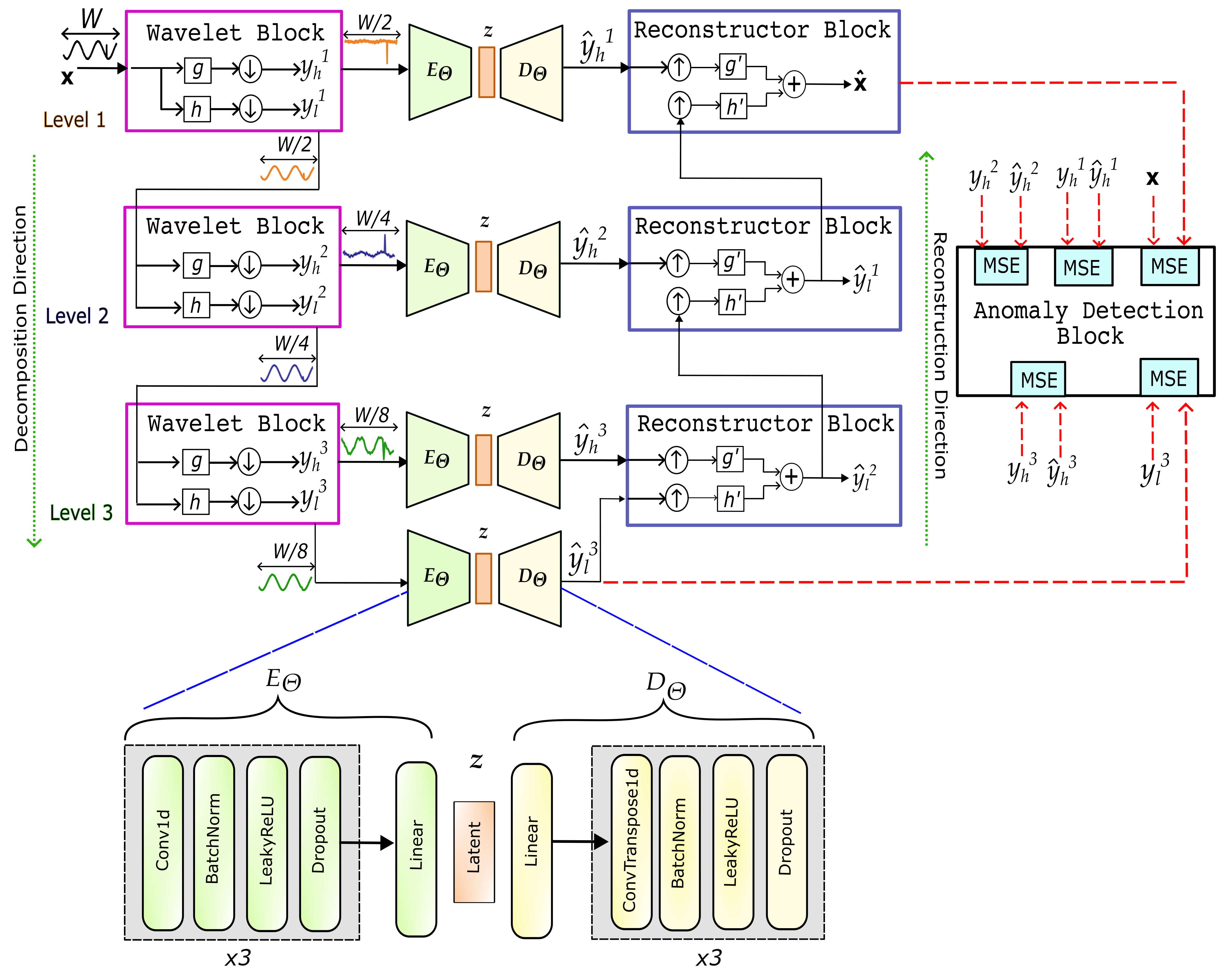}
    \caption{Overall architecture of \fname{}. It consists of 4 components: Wavelet Block, Autoencoder Network, Reconstructor Block and an Anomaly Detection Block}
    \label{fig:overview}
\end{figure*}

Figure \ref{fig:overview} provides a high-level illustration of the proposed method. The framework consists of several key components: \one A {\em Wavelet Block} which decomposes the time series into multiple frequency components. \two An
{\em Autoencoder Network} which learns the underlying features of signal and its decomposed components. \three A {\em
Reconstructor Block} which is responsible for reconstructing the signal from the learned representations. This process
is repeated across multiple levels or scales to capture both fine and coarse details. \four An {\em Anomaly Detection
Block} which evaluates the reconstructed signal using the mean-squared error (MSE) and determines whether the input
sequence is anomalous or normal. The following sections describe each component in detail.

\subsection{Wavelet Decomposition Block} 
The input time series  {\em x[n]} represented by $\textbf{x}$ is decomposed into low and high-frequency components using
the Wavelet Block of Figure \ref{fig:overview}. \emph{h} represents {\em h[n]} and \emph{g} represents {\em g[n]} for
the low-pass and high-pass filter respectively (refer to section \ref{sec:dwt}). The input signal \textbf{x} is passed
through low-pass filter  and high-pass filter separately, each followed by a downsampling operation ($\downarrow2$). The
outputs for the first level are $\mathbf{y_h^1}$ and $\mathbf{y_l^1}$ which are the \emph{detailed} and
\emph{approximation} coefficients for level-1. The approximation output $\mathbf{y_l^1}$ is next used as input to
decompose into level-2 coefficients and so on until the last level is reached. Here, L=3 and the direction of
decomposition is shown in the figure. At each level, the length of the output coefficients is halved due to the
downsampling operation. At the last level, the length of the coefficients will be $\frac{W}{2^L}$. The final output
coefficients of interest are ($\mathbf{y_l^3}$,$\mathbf{y_h^3}$, $\mathbf{y_h^2}$, $\mathbf{y_h^1}$). These are used as
inputs to a multi-level autoencoder network so that hierarchical patterns from fine to coarse scales are captured, as it was motivated in Figure \ref{fig:anomalies}(g). 

\subsection{Multi-Scale Autoencoder Network}
At each level $k$, the high-frequency component $y_h^k$ is passed through a level-specific or scale-specific autoencoder
$AE_k$ consisting of an encoder-decoder pair: $y_h^k \xrightarrow{Encoder} z^k \xrightarrow{Decoder} \hat{y}_h^k$ where
$z^k$ is the latent space encoding of $y_h^k$ and $\hat{y}_h^k $ is the output of the decoder or the reconstructed
high-frequency part. The low-frequency part $y_l^k $ is used for decomposition at the next level, except the last level,
where it is also passed through an autoencoder $AE_L$. This is done so that we can get the last level decoded $(l,h)$
pairs, which can be used by the reconstruction module for the application of inverse DWT. The reason for not passing
intermediate approximation coefficients through the autoencoder is because the final approximation coefficient $y_l^3$
already contains all low-frequency information from previous levels, making intermediate approximations ($y_l^1, y_l^2$)
redundant. Additionally, introducing more autoencoder instances would expand the parameter space, which we aim to keep
constrained.  In this way anomalies that exist at different temporal resolutions can be captured at appropriate scales
while keeping the model lightweight. 

The encoder ($E_\theta$) consists of three 1D Convolutional Layers each followed by batch normalization, LeakyReLU
activation, and dropout, progressively reducing the sequence length while increasing feature dimensions (refer to Fig
\ref{fig:overview}). We employ 1D-CNNs within the encoder-decoder architecture due to their strong ability to capture local temporal patterns in time-series data. Unlike 2D-CNNs used for image analysis, 1D-CNNs apply convolutional filters across the temporal axis of the input signal, making them well-suited for sequential data.  Additionally, compared to RNNs or LSTMs, 1D-CNNs are computationally efficient and parallelizable. 
Formally, given an input feature map $x_{in}^{(m)}$ at layer $m$ each convolutional layer
transforms it as:

\begin{equation} 
x_{\text{out}}^{(m+1)} = \text{Dropout} \Big( \text{LeakyReLU} \big( \text{BNorm}
\\
(W^{(m)} \ast x_{\text{in}}^{(m)} + b^{(m)}) \big) \Big) \nonumber
\end{equation}

where $W^{(m)}$ and $b^{(m)}$ are the convolutional kernel weights and biases at layer $m$. After $M$ convolutional
layers, the final output $x_{out}^M$ is flattened and passed through a fully connected layer:

\begin{equation}
    z = W_p \cdot \text{Flatten}(x_{out}^{(M)} + b_p) \nonumber
\end{equation}

where $W_p$ and $b_p$ are the weights and bias of the projection layer. The decoder ($D_\theta$) mirrors the encoder to
reconstruct the time series from the latent representation. It has a fully connected layer that expands the latent
vector back to a structured representation and uses three 1D Transposed Convolutional Layers which progressively
upsample the feature maps, restoring the original sequence length.

To accommodate the varying temporal resolutions, four distinct encoder-decoder pairs of the same architecture are
created, each tailored to one of the four outputs from the wavelet block ($\mathbf{y_l^3}$,$\mathbf{y_h^3}$,
$\mathbf{y_h^2}$, $\mathbf{y_h^1}$). The input sequence length for each encoder-decoder pair is carefully adjusted to
match the scale of the wavelet components they process. The first encoder-decoder pair, responsible for $y_h^1$
operates on an input sequence of length $\frac{W}{2}$. Similarly, at the coarsest level, the encoder-decoder pair for
$(y_h^3,y_l^3)$ operates on an input sequence of length $\frac{W}{2^3}$.  Finally, the 4 inputs and outputs of the
autoencoders  are retained to be fed into an anomaly detection module before going through the reconstruction module. 

As shown in Figure 1(a), using a standard autoencoder directly on the raw time series, without multi-scale decomposition, failed to detect subtle anomalies. The simple AE in that figure received only the original time-series input and reconstructed it at a single scale. Since it lacked frequency separation or hierarchical feature learning, the subtle deviation was not captured, resulting in a low reconstruction error. This limitation motivated our design of a scale-aware architecture which takes information from distinct frequency bands into account.  

\subsection{Reconstruction Block}
\label{sec:reconstruction}
The reconstruction module applies the inverse DWT (refer to section \ref{sec:dwt}) starting from the last level to
reconstruct the input as follows: The last level's $(\hat{y}_l^L,\hat{y}_h^L)$ decoded components are used by the
reconstruction module to get the low-frequency part generated by the previous level: $\hat{y}_l^{L-1} =
IDWT(\hat{y}_l^L,\hat{y}_h^L)$ where IDWT is defined using Equation ~\eqref{eq:idwt}. Since we already have
$\hat{y}_h^k$ for all $k$ levels, this process is repeated iteratively until reaching level $1$, where we obtain the
final reconstructed signal: $\hat{x} = IDWT(\hat{y}_l^1,\hat{y}_h^1)$.

\subsection{Anomaly Detection Module}
In a standard autoencoder, anomaly detection (\cite{anomaly-review}) is typically performed by computing the L2-Norm
between the original and reconstructed signal ($x,\hat{x}$) as per Equation ~\eqref{eq:error}. However, since our
approach operates at multiple scales, we compute reconstruction errors at each level to capture anomalies across
different frequency bands. Specifically, the high-frequency reconstruction error at each level $k$:

\begin{equation}
    \mathcal{L}_h^k = ||y_h^k-\hat{y}_h^k||_2
    \label{eq:err_h}
\end{equation}
The low-frequency reconstruction error at the final level $L$ is given by,

\begin{equation}
    \mathcal{L}_l^L = ||y_l^L-\hat{y}_l^L||_2
    \label{eq:err_l}
\end{equation}

The total anomaly score is then determined as a weighted sum of these errors, incorporating both multi-scale
high-frequency errors and the final low-frequency error, along with the weighted reconstruction error of the original
signal:

\begin{equation}
    \mathcal{L}_{total} = \beta||x-\hat{x}||_2+ \sum_{k=1}^{L}\lambda_k \mathcal{L}_h^k + \gamma\mathcal{L}_l^L
    \label{eq:err_total}
\end{equation}

where $\beta$,$\lambda$ and $\gamma$ are the weights for different scales.  Utilizing this multi-scale reconstruction
error ensures that anomalies across all scales are accounted for in the detection process through a weighted sum. While
some anomalies like abrupt spikes, may only show up at high frequencies, others, such as long-term drifts, may show up
at low frequencies. By aggregating the reconstruction errors across multiple scales, even subtle anomalies that significantly affect only a single scale will yield a large error in that term. When summed, this leads to a more prominent total anomaly score, enhancing the model’s ability to detect subtle or localized anomalies. Furthermore, potential false positives arising from noise in specific frequency bands can be mitigated by appropriately adjusting the weights $\lambda_k$ and $\gamma$.
Finally, the anomaly score from Equation
~\eqref{eq:err_total} is computed for each window of the time series. This score across the window is compared against a
threshold $\tau$ and if it exceeds $\tau$ the window is flagged as anomalous. Notably,$ \mathcal{L}_{total} $ is not only used as an anomaly scoring function but is also employed as the loss function during training. This design choice ensures that the autoencoder is explicitly optimized to reduce reconstruction discrepancies across multiple wavelet subbands, thereby aligning the training objective with the detection criterion.

\section{Experimental Settings}
\label{sec:experiments}
\subsection{Datasets}
\subsubsection{Numenta Anomaly Benchmark Dataset}
NAB is a benchmark for detecting anomalies in streaming real-time applications (\cite{b2}). It consists of 58 timeseries data with identified abnormal periods of behavior from both artificial and real-world sources including traffic data, AWS server metrics, Twitter volume, and metrics related to advertisement clicking. The datasets used for our experiments belong to the category for which the causes of the anomalies are known without any hand labeling (\cite{tsvae}). These include:
\begin{itemize}
\item \emph{\bfseries NYC Taxi: } Represents a time series for the number of NYC taxi passengers. There are five anomalies that occur during the NYC marathon, Thanksgiving, Christmas, New Years day, and a snow storm.
\item \emph{\bfseries Ambient Temperature: } The ambient temperature in an office.
\item \emph{\bfseries CPU Utilization: } Captures CPU utilization data from Amazon Web Services (AWS).
\item \emph{\bfseries EC2 Request Latency: }Contains CPU usage data from a server in AWS’s East Coast data center.
\item \emph{\bfseries Machine Temperature: } Monitors the internal temperature of an industrial machine component. 
\end{itemize}
\subsubsection{Yahoo Webscope S5 Dataset} 
The Yahoo (\cite{yahoo}) dataset comprises four sub-benchmarks for anomaly detection. These are named as \emph{\bfseries A1, A2, A3} and \emph{\bfseries A4}. These contain either real or synthetic web traffic metrics with labels. \emph{A1 Benchmark} consists of real traffic data with point and collective anomalies while the other three benchmarks are synthetic in nature. We have used A1 and A2 Benchmarks for our experiments, as they are univariate datasets, whereas A3 and A4 are multivariate (\cite{cvae}).

\subsection{Training }
The model is trained using only normal samples (\cite{anomaly-review}), such that deviations from the learned patterns (i.e., anomalies) result in higher reconstruction errors during inference. Training and testing data are divided so that, for all datasets, normal samples are used for training, while anomalous samples are reserved for testing (\cite{tsvae}). In case there are anomalies in the training set, we fill the anomalies with zero as done by (\cite{fcvae}). 
Training is carried out setting a learning rate of $10^{-3}$, Adam as optimizer, and a weight decay of $10^{-6}$. These are selected on the basis of preliminary experiments for stable convergence and improved generalization. Haar wavelet function (with $L=3$) is used throughout the experiments for the wavelet block due to its simplicity and favorable trade off between higher-order representations and model complexity. The loss function is the same as defined in equation~\eqref{eq:err_total}. A key aspect of training is tuning the loss function parameters in the final objective function, $\mathcal{L}_{total}$. These parameters regulate the weighting of data from various frequency levels in the loss computation. Proper tuning is essential to guarantee that the model captures both local changes and long-term dependencies since wavelet decomposition yields both low-frequency (trend-related) and high-frequency (detail-related) components. Depending on the distinct time series properties of each dataset, different weight combinations may be needed, so we train different models for each dataset's curve within a benchmark. We have conducted an extensive grid search to determine the optimal weight combinations for each dataset. All the experiments are implemented in PyTorch and conducted on a single NVIDIA RTX A6000 GPU.

\subsection{Evaluation}
During testing phase, data containing anomalies is used for evaluation. We use {\em Precision}, {\em Recall} and {\em F1-Score} as evaluation metrics:
\begin{equation}
    P = \frac{TP}{TP+FP}, R=\frac{TP}{TP+FN}, F1=\frac{2 \times P \times R}{P+R} \nonumber
    \label{p}
\end{equation}
where P is the precision, R is the recall, F1 is the F1-Score, TP is true positive (correctly detected anomalies), FP is false positive (normal points mistakenly classified as anomalies) and FN is false negative (anomalies that were not detected). However, in real-world anomaly detection, the goal is not just to pinpoint the exact moment an anomaly occurs but rather to identify a broader window of anomalous activity where an alert can be triggered. This aligns with practical scenarios where timely detection is more critical than pinpoint accuracy. To address this we have used the adjustments from (\cite{tsvae}) following (\cite{eval}) which essentially says that the entire anomaly segment will be regarded as anomalous even if one point in the segment is flagged as an anomaly. Since our method utilizes overlapping sliding windows for anomaly detection, a single anomaly point may appear in multiple windows as they shift over time and without proper adjustment, this could lead to an inflated false positive rate. The threshold $\tau$ for anomaly detection is manually chosen by evaluating the F1 score over a predefined range of thresholds and selecting the one that produces the highest F1 score. We further report the latency and power consumption  of the proposed method on an edge-AI platform \emph{NVIDIA Jetson Nano} (\cite{jetson}) to reflect the ease of deployment of the model on edge devices.

\section{Results and Analysis}
\label{sec:results}
\subsection{Quantitative Results}
\begin{table*}[h!]
\centering
\caption{Performance comparison on Yahoo and NAB datasets. 
NT - NYC Taxi, ER - EC2 Request, CU - CPU Utilisation, AT - Ambient Temperature, MT - Machine Temperature}

\renewcommand{\arraystretch}{1.4}
\setlength{\tabcolsep}{9pt}

\begin{tabular}{c c cc ccccc}
\toprule

\multirow{2}{*}{\textbf{Metrics}} & \multirow{2}{*}{\textbf{Method}} 
& \multicolumn{2}{c}{\textbf{Yahoo}} 
& \multicolumn{5}{c}{\textbf{NAB} (\emph{realKnownCause})} \\

\cmidrule(lr){3-4} \cmidrule(lr){5-9}

& & \textbf{A1} & \textbf{A2} 
& \textbf{NT} & \textbf{ER} & \textbf{CU} & \textbf{AT} & \textbf{MT} \\

\midrule

\multirow{4}{*}{\textbf{F1-Score}} 
& \textbf{\fname{}} & \textbf{0.942} & \textbf{0.986} & \textbf{0.953} & \underline{\textbf{0.998}} & \textbf{0.975} & \textbf{1.0} & \underline{\textbf{0.600}} \\
& \textbf{LSTM-VAE} & \underline{\textbf{0.919}} & 0.644 & 0.570 & 0.994 & 0.691 & \underline{\textbf{1.0}} & \underline{\textbf{0.748}} \\ 
& \textbf{TADGAN} & 0.618 & 0.803  & 0.750 & 1.0 & 0.500 & 0.667 & - \\
& \textbf{FCVAE} & 0.791 & \underline{\textbf{0.895}}  & \underline{\textbf{0.860}} & \underline{\textbf{1.0}} & \underline{\textbf{0.827}} & 0.667 & 0.2 \\ 

\midrule

\multirow{4}{*}{\textbf{Recall}} 
& \textbf{\fname{}} & 0.934 & \textbf{1.0} & \textbf{1.0} & \textbf{1.0} & \textbf{1.0} & \textbf{1.0} & \textbf{1.0}  \\
& \textbf{LSTM-VAE} & \textbf{0.980} & 1.0 & 0.4 & 1.0 & 1.0 & 1.0 & 1.0 \\ 
& \textbf{TADGAN} & 0.606 & 0.925 & 0.600 & 1.0 & 1.0 & 1.0 & - \\
& \textbf{FCVAE} & 0.975 & 0.9872 & 0.800 & 1.0 & 1.0 & 0.500 & 1.0 \\

\midrule

\multirow{4}{*}{\textbf{Precision}} 
& \textbf{\fname{}} & \textbf{0.970} & \textbf{0.979} & 0.911 & \textbf{0.996} & \textbf{0.951} & \textbf{1.0} & 0.426 \\
& \textbf{LSTM-VAE} & 0.915 & 0.492 & 1.0 & 0.989 & 0.528 & 1.0 & \textbf{0.598} \\ 
& \textbf{TADGAN} & 0.716 & 0.760 & 1.0 & 1.0 & 0.330 & 0.5 & - \\
& \textbf{FCVAE} & 0.784 & 0.867 & \textbf{0.932} & 1.0 & 0.705 & 1.0 & 0.12 \\

\bottomrule
\end{tabular}

\label{tab:performance}
\end{table*}
\begin{table}
    \centering
    \caption{Window Length settings for different methods}
    \renewcommand{\arraystretch}{1.4} 
    \setlength{\tabcolsep}{8pt} 
    \begin{tabular}{c c c}
    \hline
    % \rowcolor{lightgray}
    \textbf{Method} & \textbf{Window Length (W)} & \textbf{Dataset} \\
    \hline
    \fname{} & 64 & All datasets \\
    \hline
    VAE-LSTM & 168 & Yahoo \\
    % \cline{2-3}
    & 168 & NT \\ 
    % \cline{2-3}
    & 192 & ER \\ 
    % \cline{2-3}
    & 144 & CU \\
    % \cline{2-3}
    & 168 & AT \\
    % \cline{2-3}
    & 288 & MT  \\
    \hline
    FCVAE & 48 & All datasets \\
    \hline
    TADGAN & 100 & All datasets \\
    \hline

    \end{tabular}
    \label{tab:W}
\end{table} 
Table \ref{tab:performance} presents the performance comparison of our method with the baseline models. \fname{} achieves the highest F1 Score on the Yahoo Benchmark and outperforms the latest competitor by $19\%$ and $10\%$ for the $A1$ and $A2$ Benchmark, respectively. For the NAB Benchmark, \fname{} outperforms NT by $10.8\%$, CU by $17.89\%$, AT by $50\%$. For ER, the F1 Score remains competitive, closely matching the best-performing models. This demonstrates that our lightweight model can effectively detect anomalies while competing with more parameter-intensive approaches. 

Although VAE-LSTM (\cite{tsvae}) has good recall across most datasets, it relies on a significantly larger window length than our model. Table \ref{tab:W} lists the window length values for all datasets and methods, highlighting that VAE-LSTM consistently uses larger windows. This is due to its hybrid architecture, where the VAE model is trained on overlapping sliding windows of length $p$, and the LSTM model processes a non-overlapping sequence of length $k$ from the VAE embeddings. During the evaluation, the model analyzes a test sequence spanning $p \times k$ readings, leading to an effectively larger window size than ours. TADGAN (\cite{tadgan}) also uses a considerably larger window size to detect anomalies. FCVAE (\cite{fcvae}) takes a single-model approach, where one model is trained for all dataset curves within a benchmark. While this strategy offers generalizability, it may struggle to capture dataset-specific variations effectively. In contrast, we train separate models for each dataset curve, allowing us to fine-tune the weights ($\beta$, $\lambda$, $\gamma$) based on the unique characteristics and patterns of each dataset.

\subsection{Qualitative Results}
\label{sec:qual}
\begin{figure*}
    \centering
    \includegraphics[width=1\linewidth]{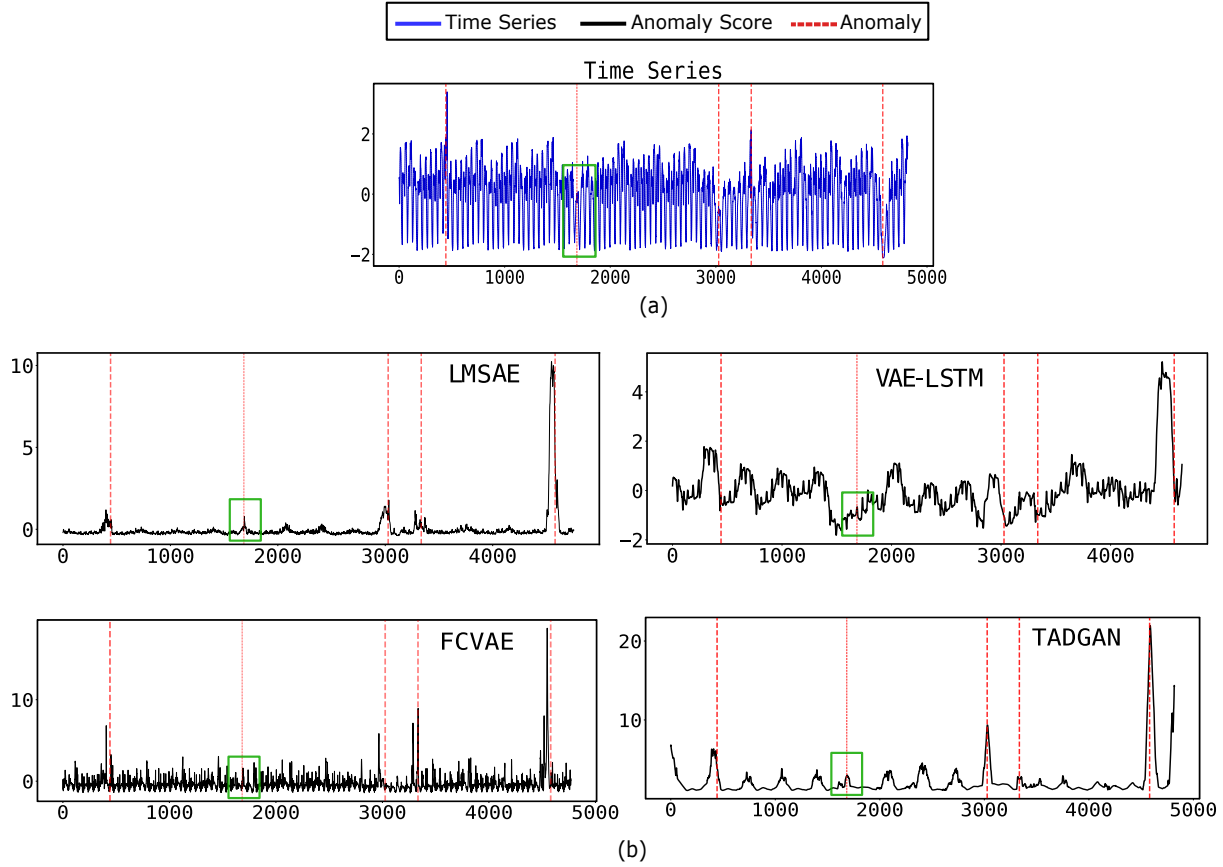}
    \caption{ (a) represents the time series input (test data) and (b) represents Anomaly Scores of different methods for \emph{NYC Taxi (NT)} of \emph{NAB Benchmark}. The green box shows the anomaly under consideration.}
    \label{fig:results_nyc}
\end{figure*}
\begin{figure*}
    \centering
    \includegraphics[width=1\linewidth]{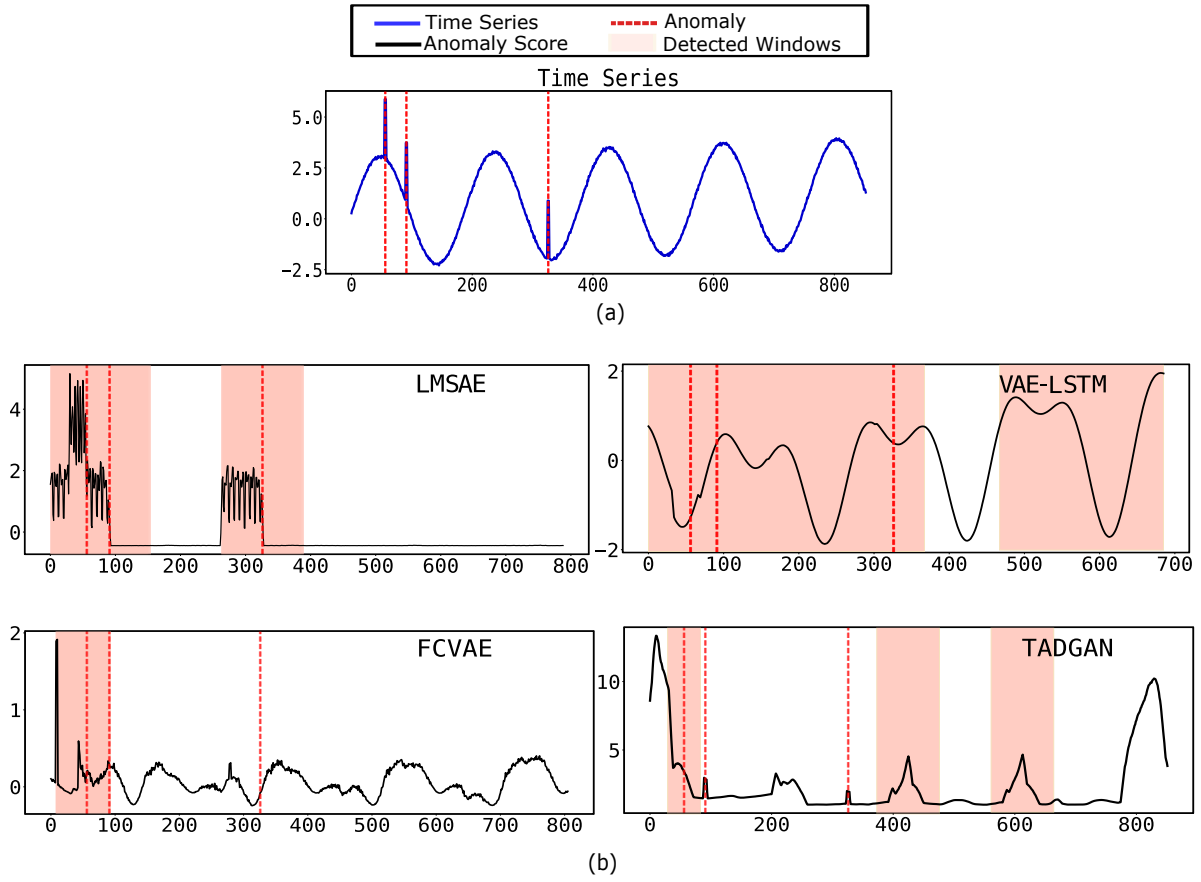}
    \caption{Anomaly Scores of different methods for a curve of \emph{A2 Benchmark}. (a) shows the time series test data and (b) shows the Anomaly Scores with detected windows highlighted.}
    \label{fig:results_a2}
\end{figure*}
The anomaly scores produced by different detection techniques on \emph{NYC Taxi} Dataset (\emph{NAB Benchmark}) are displayed in Figure \ref{fig:results_nyc}. 
It is clear that most of the baseline approaches produce very smooth or ambiguous anomaly scores for the anomaly section outlined within the green box, making it challenging to discern abnormal behavior from normal fluctuations. FCVAE, utilizing FFT and attention mechanisms to integrate both global and local frequency features, may miss subtle, transient anomalies, particularly those that only affect the most recent time point, due to its dependence on globally averaged frequency representations. In contrast, our method produces a sharp and well-localized spike in the anomaly score, indicating its ability to detect even subtle anomalies. This improved sensitivity can be attributed to the effective weighting of time and frequency domain features, allowing the model to better discriminate between minor deviations and normal variability.

In {\em A2 Benchmark}, there is random seasonality, trend, and noise present in the time series curves, which makes it difficult to flag an anomaly against a threshold as the anomaly scores are affected by it. One of the example from \emph{A2Benchmark} is shown in Figure \ref{fig:results_a2}. 
Our model is able to perform the best on {\em A2 Benchmark} because we can choose to use the weight values of the detail coefficients only, since the trend part of the signal primarily resides in the original and approximation part. By setting $\beta$ and $\gamma$ to zero, we pay attention to the high frequency details only, which suppresses low-frequency trends and enhances sensitivity to abrupt changes given by the detail coefficients. The anomaly score for the anomalous windows for our model is different from the anomaly score of normal samples which is not that easy to distinguish from other methods and hence leads to missing the true anomalies. 

\subsection{Latency and Power Consumption on NVIDIA Jetson Nano} 
%%%%%%%%
\begin{table}[h!]
    \centering
    \caption{Jetson Nano Platform Configuration}
    \renewcommand{\arraystretch}{1.4} 
    \setlength{\tabcolsep}{9pt} 
    \begin{tabular}{c c}
    \hline
    % \rowcolor{lightgray}
    \textbf{Parameter} & \textbf{Type/Value} \\
    \hline
    CPU & ARM Cortex-A57 \\
    $\#$CPU Cores & 4 \\
    CPU Frequency (Max) & 1.43GHz \\
    \hline
    GPU & NVIDIA Maxwell Architecture \\
    $\#$CUDA Cores & 128 \\
    GPU Frequency (Max) & 921MHz \\
    \hline
    RAM & 4GB 64-bit LPDDR4 \\
    \hline
    Power & 5W-10W \\
    \hline 
    \end{tabular}
    \label{tab:jetson}
\end{table}
%%%%%%%%%%%%%

To evaluate the suitability of our proposed model for resource-constrained environments, we conducted experiments on the NVIDIA Jetson Nano, a low-power, edge AI platform designed for running machine learning models on-device. It combines low-power GPUs with ARM-based CPUs in a compact form factor, offering a highly efficient performance-to-energy ratio ideal for edge AI applications. The configuration details of the platform are given in Table \ref{tab:jetson}.

We used NVIDIA TensorRT, a C++ library that enables efficient and accelerated inference on NVIDIA GPUs. It takes a trained model, defined by its architecture and learned weights, and converts it into a runtime optimized engine. To maximize performance, TensorRT applies techniques like layer fusion, graph optimizations, and selects the most efficient kernel implementations from its library of optimized operations (\cite{deep-edge-bench}). 

\subsubsection{Runtime Latency Results}
\begin{figure}
    \centering
    \includegraphics[width=0.7\linewidth]{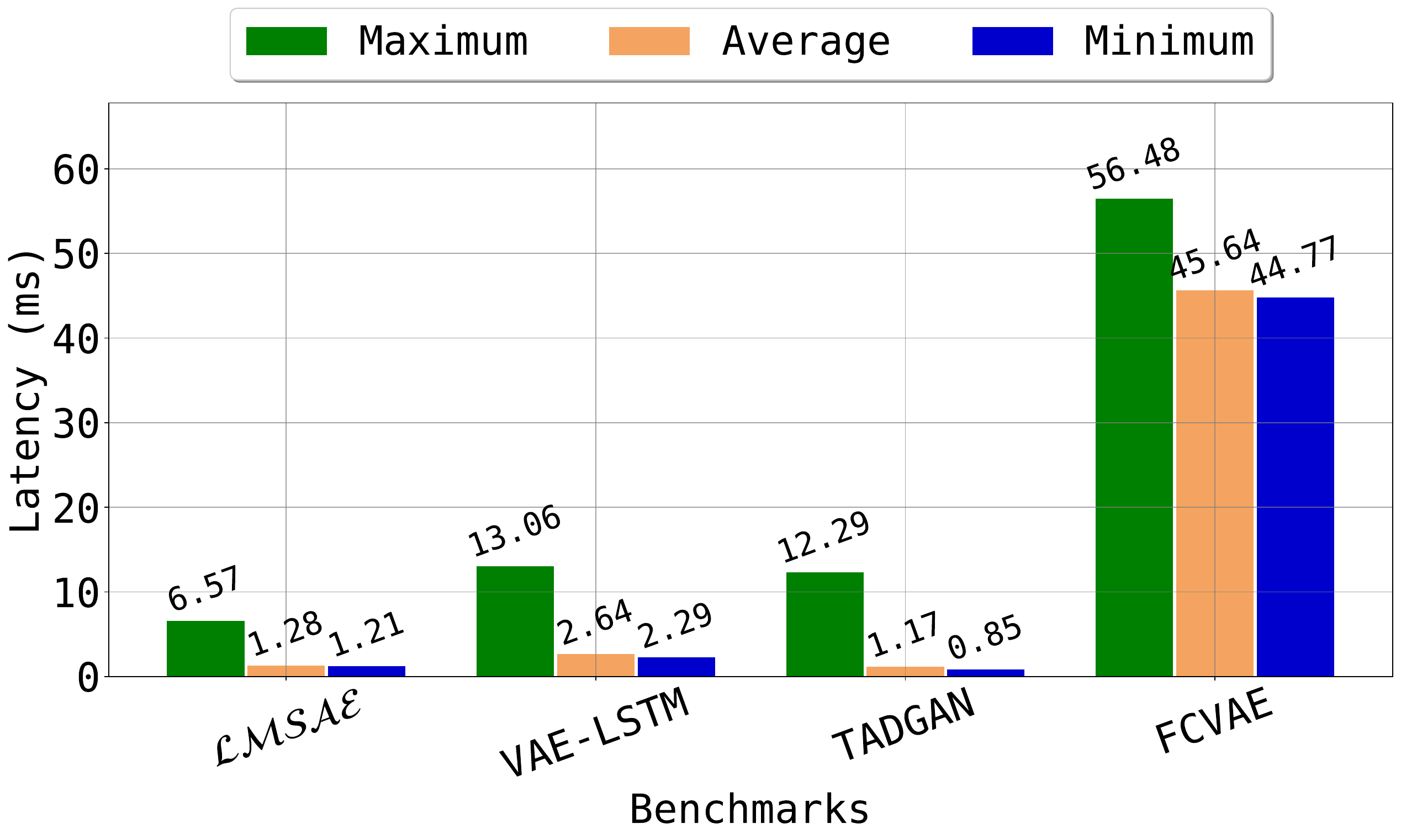}
    \caption{Inference latency}
    \label{fig:latency}
\end{figure}
Figure \ref{fig:latency} presents the latency measurements of different methods. The reported latency includes the time taken for host-to-device (H2D) input data transfers, device-to-host (D2H) output data transfers, and the GPU computation time. It is averaged for 10 inferences. From the figure, it is evident that our proposed model exhibits the lowest maximum latency among the other methods. In contrast, FCVAE has the highest latency (max, min and average), primarily due to its larger parameter space and associated computational complexity. Although TADGAN and VAE-LSTM achieve somewhat competitive minimum latencies, their maximum latencies remain significantly higher than ours, indicating they are less stable under different loads. 

\subsubsection{Runtime Power Consumption Results}
\begin{figure}
    \centering
    \includegraphics[width=0.7\linewidth]{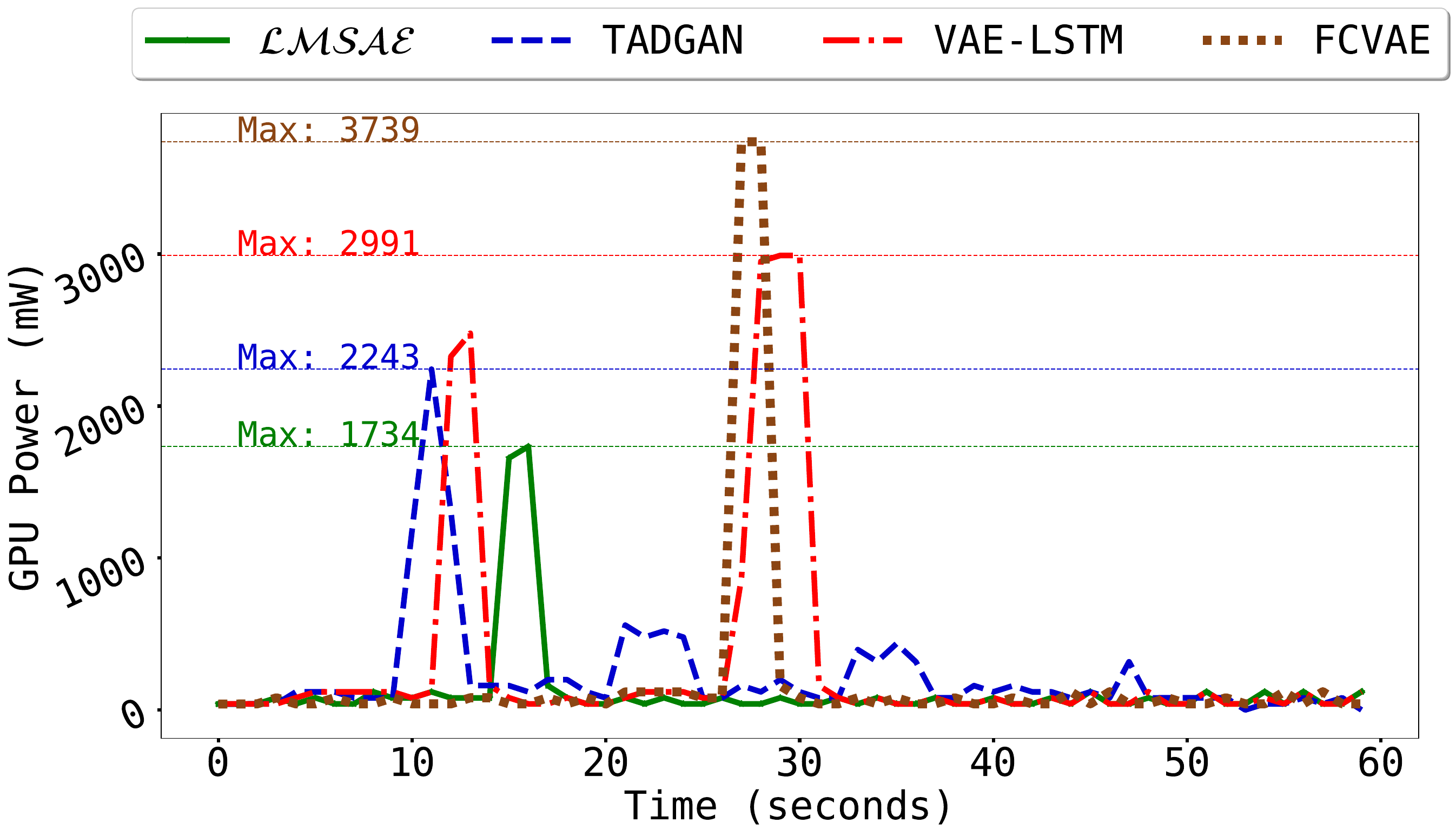}
    \caption{Power consumption}
    \label{fig:power}
\end{figure}

For power consumption, there are three sensors which are located at the power input of the board, CPU and GPU. These sensors can be read to monitor power consumption. Figure \ref{fig:power} shows the GPU power consumption. The peak power consumed during inference is the least for our model. FCVAE, while already showing high latency, also shows the highest power peaks, suggesting that it is computationally and energy inefficient compared to lighter models like ours.

\section{Model Interpretation}
\label{sec:interpretation}
To gain deeper insights into the behavior of our multi-scale autoencoder, we examine \one how our autoencoder reconstructs the input signals in the presence of anomalies and \two how the detection performance varies with the parameters of our multi-scale loss function given by Equation \eqref{eq:err_total}. 

\subsection{Reconstruction of the Original Signal and Frequency Components}
\begin{figure*}
    \centering   \includegraphics[width=1\linewidth]{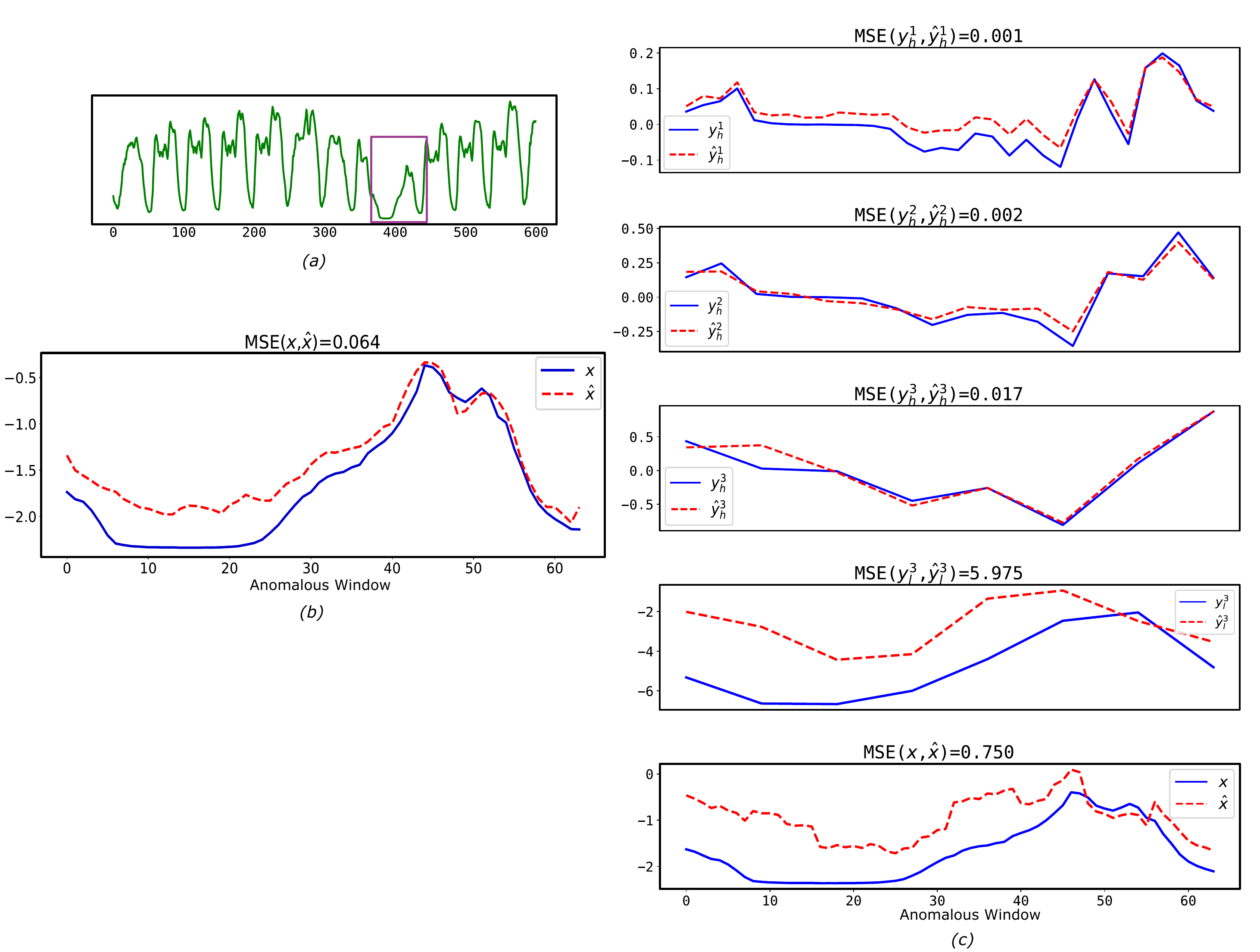}
    \caption{\emph{(a)} represents the input time series with the purple box highlighting the anomalous window. \emph{(b)} shows the reconstruction comparison between an autoencoder lacking multi-scale view and \emph{(c)} shows the same for our proposed multi-scale autoencoder \fname{}. Blue solid lines represent original series whereas red dashed lines represent reconstructed part.}
    \label{fig:recon_err}
\end{figure*}

Figures \ref{fig:recon_err}(b) and \ref{fig:recon_err}(c) provide a visual comparison of reconstruction errors for an anomalous window of the \emph{NYC Taxi} dataset \ref{fig:recon_err}(a). The anomalous window is fed as input to our multi-scale autoencoder, which processes it through a discrete wavelet transform (DWT) block to obtain multi-resolution components: $y_h^1$, $y_h^2$, $y_h^3$, and $y_l^3$. These components are encoded and decoded to produce corresponding reconstructions: $\hat{y}_h^1$, $\hat{y}_h^2$, $\hat{y}_h^3$, and $\hat{y}_l^3$. We compare each of these reconstructed components to their ground truth DWT counterparts to assess subband-wise reconstruction accuracy and observe in which frequency band this anomaly might manifest. The final reconstructed signal $\hat{x}$ is then obtained by applying the inverse wavelet transform to the decoded coefficients. To contrast the performance of our multi-scale autoencoder, we feed the same anomalous window to a simple autoencoder without multi-scale view. Unlike a simple autoencoder that estimates \( \hat{x} \) in a flat fashion, the structure of \fname{} allows reconstruction errors from different subbands and resolutions to be faithfully represented in the final loss. 
In Figure \ref{fig:recon_err}(b), the standard autoencoder, which lacks a multi-scale perspective, generates a reconstructed signal that closely mirrors the original signal (MSE of $0.06$), despite the presence of an anomaly. The reconstruction error which is the difference between the original signal and its reconstruction, is notably small. This poses a significant challenge for anomaly detection:
\one \emph{Subtle Errors}: A small reconstruction error implies that the autoencoder struggles to differentiate anomalous features from normal patterns.\two \emph{Detection Difficulty}: With such minimal deviation, distinguishing the anomalous window from typical data becomes nearly impossible, as the error is too subtle to trigger detection thresholds. \three \emph{Practical Impact}: In real-world scenarios, this can result in missed anomalies, undermining the reliability of systems dependent on accurate flagging, such as fault detection or security monitoring.

Conversely, Figure \ref{fig:recon_err}(c) showcases the multi-scale autoencoder's superior performance. This approach reveals a larger reconstruction error, particularly evident in the coarse-level component ($y_l^3$), which represents the signal’s low-frequency, overall structure. Following key insights can be derived:
\one \emph{Enhanced Sensitivity}: The pronounced error in $y_l^3$ (MSE of $5.975$) indicates our multi-scale autoencoder's ability to detect significant deviations caused by the anomaly, especially those affecting the signal’s broader trends for this example. 
\two \emph{Anomaly Amplification}: Unlike the standard autoencoder, the multi-scale method amplifies the anomaly's presence, with a difference of $0.75$ between the original signal $(x)$ and reconstructed signal $(x')$ which is much greater than $0.06$ for the simple autoencoder \three \emph{Threshold Simplicity}: This larger error facilitates setting a clear detection threshold. Simply observing the pair $(x, x')$ can suffice to flag anomalies in this case, streamlining the detection process.

\begin{figure*}
    \centering   \includegraphics[width=1\linewidth]{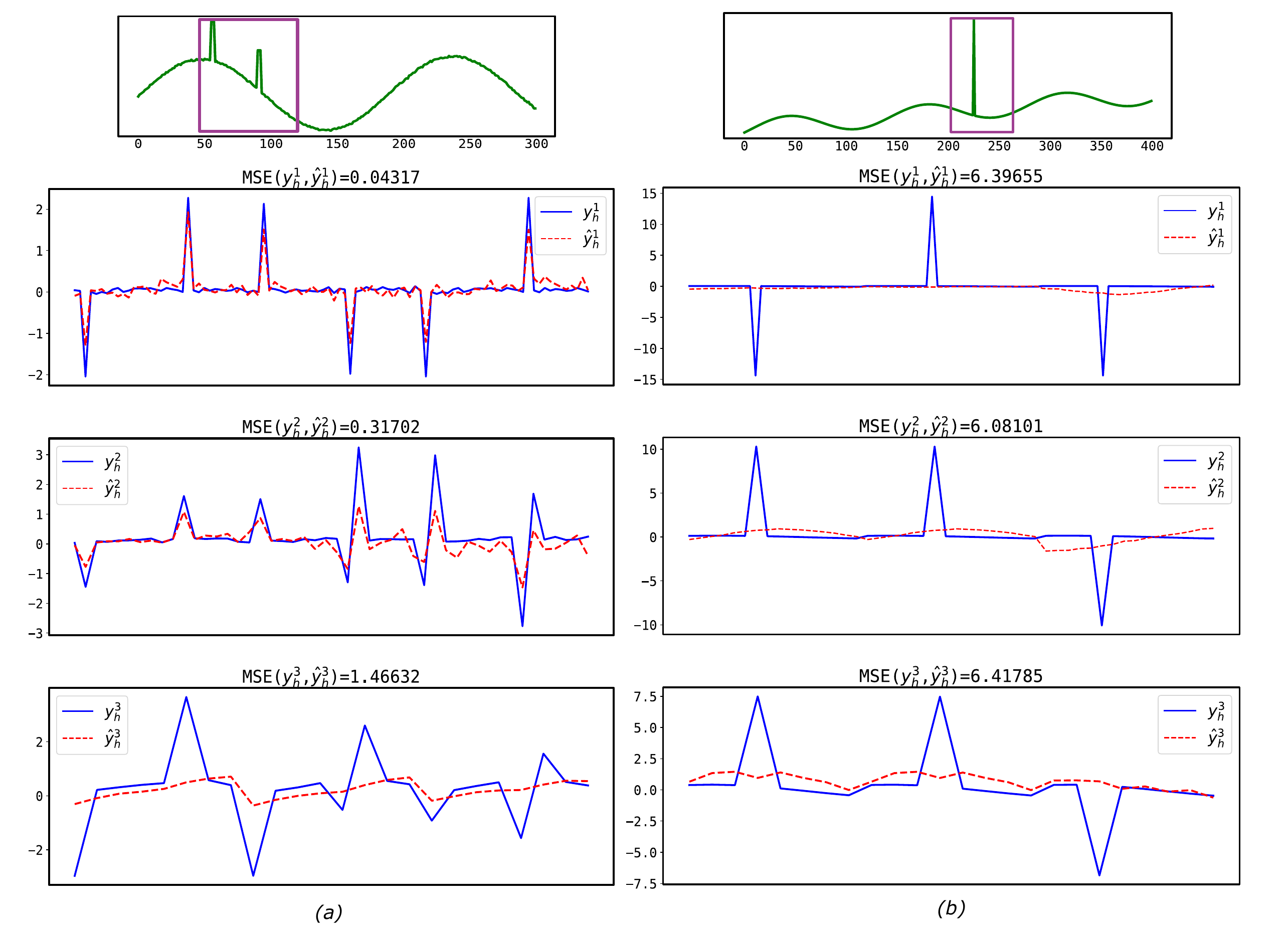}
    \caption{Sub-band reconstruction comparisons for two different synthetic anomalous signals from the \emph{A2 benchmark} using our multi-scale autoencoder \fname{}. The plots at the top represent the input time series with an anomaly segment highlighted in the purple box. Blue solid lines represent original series whereas red dashed lines represent reconstructed part.}
    \label{fig:recon_err_a2}
\end{figure*}
We further analyze how anomalies manifest across different high frequency sub-bands using controlled synthetic anomalies from the \emph{A2 Benchmark}. Figures \ref{fig:recon_err_a2}(a) and \ref{fig:recon_err_a2}(b) show subband-wise reconstruction results for two distinct synthetic signals for three consecutive anomalous regions. 
In Figure \ref{fig:recon_err_a2}(a), the reconstruction errors are concentrated in the third level high-frequency component ($y_h^3$), where the model exhibits a notable mismatch between the true and reconstructed coefficients (MSE $=1.46$), while the errors in the first and second levels are relatively small. This suggests that the anomaly primarily distorts the coarse-scale high frequency features, predominantly in the lower-resolution band.
In contrast, Figure \ref{fig:recon_err_a2}(b) displays significant reconstruction errors across all three high-frequency bands ($y_h^1$, $y_h^2$, and $y_h^3$), with MSEs exceeding $6.0$ in each. This indicates that the anomaly affects both fine-grained details and global structure, producing high-frequency spikes and broader disruptions. Notably, the reconstructed coefficients (red dashed line) fail to match the sharp transitions present in the original components (blue solid line), confirming that the anomaly is detectable at all resolutions. Both of these examples are also consistent with the weighting strategy described in Section~\ref{sec:qual} for Figure~\ref{fig:results_a2}, where only the $\lambda_i$ terms were emphasized while the coefficients $\beta$ and $\gamma$ were set to zero.

The hierarchical design of \fname{}, thus, excels at capturing both fine details and structural changes, making it particularly adept at highlighting anomalies that manifest in the different frequency components of the signal. 

\subsection{Sensitivity Analysis}
In this section, we study the sensitivity of our model to the weight parameters $\beta,\lambda_i,\gamma$ given in Equation \eqref{eq:err_total} during training and testing.
 We vary each parameter from $0.1$ to $3$ with a step size of $0.1$ while keeping others fixed and plot the F1-Score for each parameter value. We do this for datasets from \emph{NAB Benchmark} (NT,CU,AT,MT).

\begin{figure}[!htb]
    \centering
    \begin{minipage}[t]{0.48\textwidth}
        \centering
        \includegraphics[width=\linewidth]{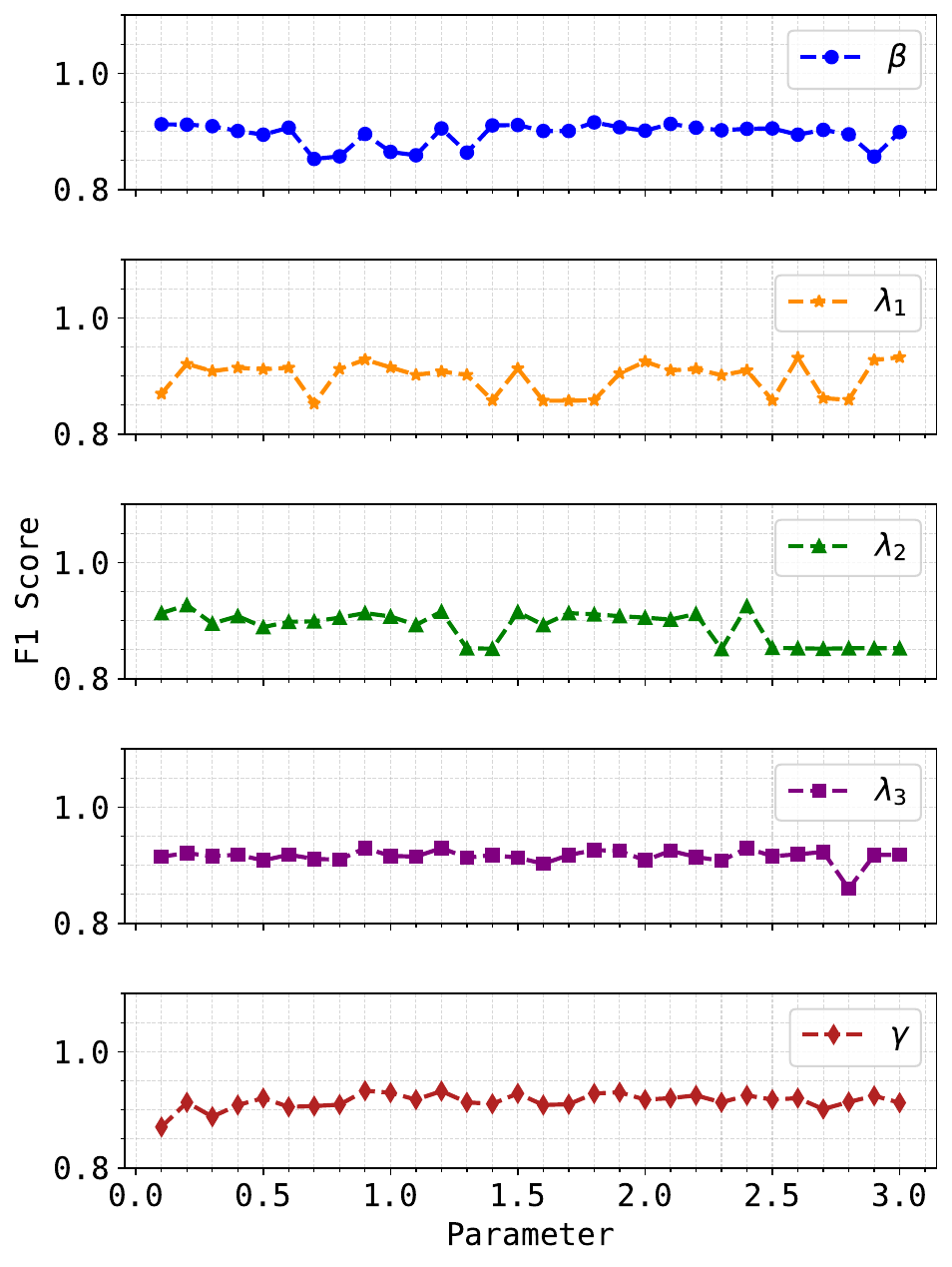}
        \caption{Sensitivity of F1-Score for given parameters for \emph{NYC Taxi (NT)} dataset.}
        \label{fig:nyc_sa}
    \end{minipage}%
    \hfill
    \begin{minipage}[t]{0.48\textwidth}
        \centering
        \includegraphics[width=\linewidth]{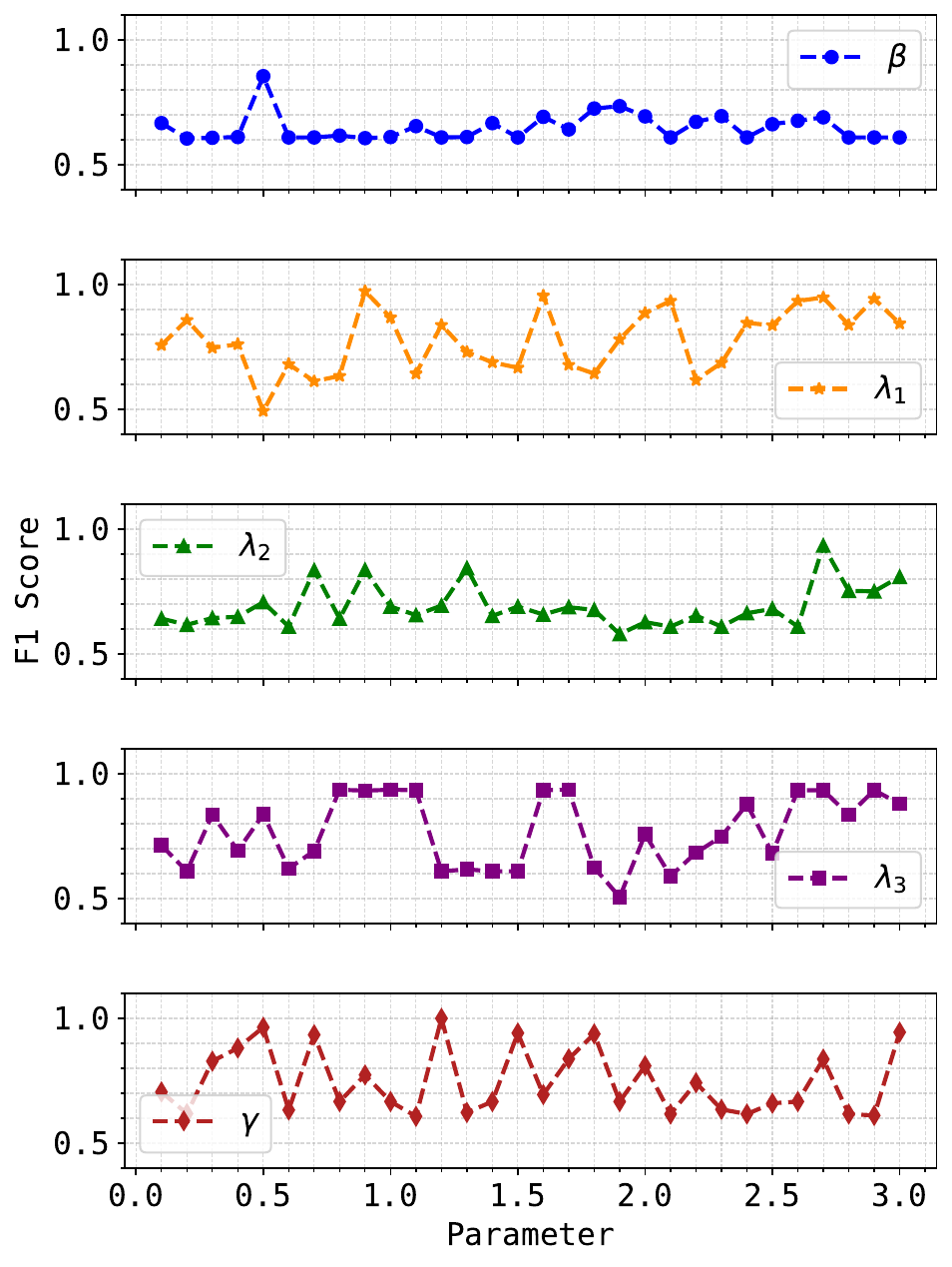}
        \caption{Sensitivity of F1-Score for given parameters for \emph{CPU Utilization (CU)} dataset.}
        \label{fig:cpu_sa}
    \end{minipage}
\end{figure}
Figure \ref{fig:nyc_sa} shows the variation of F1-Score with these parameters for the \emph{NYC Taxi} dataset. The model appears to be most sensitive to the weights associated with Level $1$ ($\lambda_1$) and Level $2$ ($\lambda_2$) detail coefficients, indicating that these parameters should be tuned more carefully than others. Additionally, while $\beta$ shows notable fluctuations up to approximately $\beta = 1.4$, its influence stabilizes afterward, with a slight drop toward the higher end of the range. The sensitivity of the model differs across datasets. Figure \ref{fig:cpu_sa} and \ref{fig:ambient_sa} shows the plots for \emph{CPU Utilization} dataset and \emph{Ambient Temperature} dataset respectively. 

For CU, we can observe that the model is mostly sensitive to $\gamma$ and $\lambda_i$'s. This means that choosing parameters for these can greatly affect our detection performance. This finding supports the idea that subtle anomalies can manifest in multiple frequency bands and cannot be effectively detected if the focus is placed solely on reconstructing the original signal (via $\beta$). For \emph{Ambient Temperature} dataset, the model is more sensitive to the weights of the detail coefficients, while $\beta$ and $\gamma$ exhibit regions where the model performance remains relatively stable.

\begin{figure}
    \centering
    \includegraphics[width=0.7\linewidth]{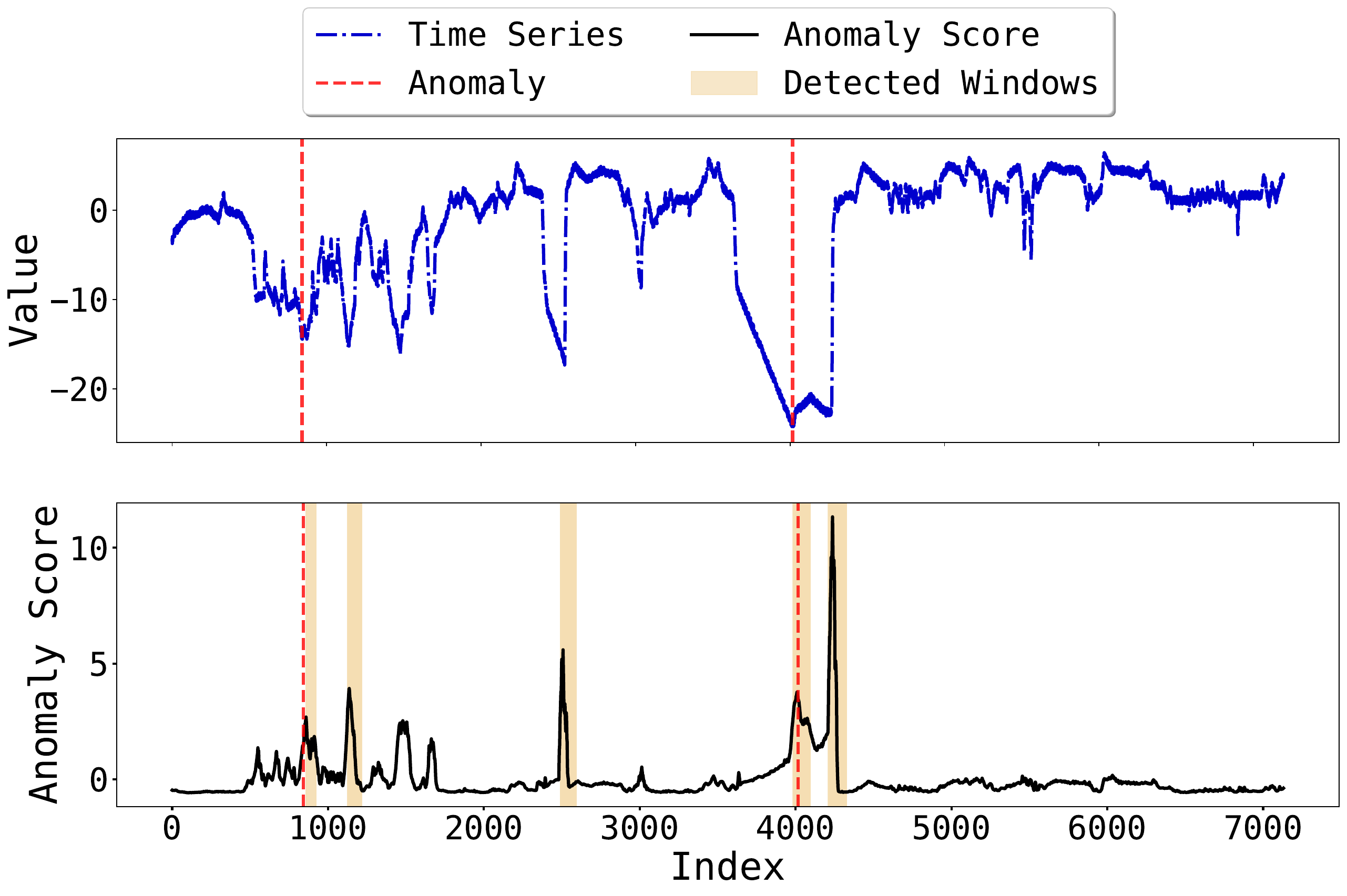}
    \caption{Machine temperature dataset with anomaly score produced by \fname{}.}
    \label{fig:machine_data}
\end{figure}

\begin{figure}[!htb]
    \centering
    \begin{minipage}[t]{0.48\textwidth}
        \centering
        \includegraphics[width=\linewidth]{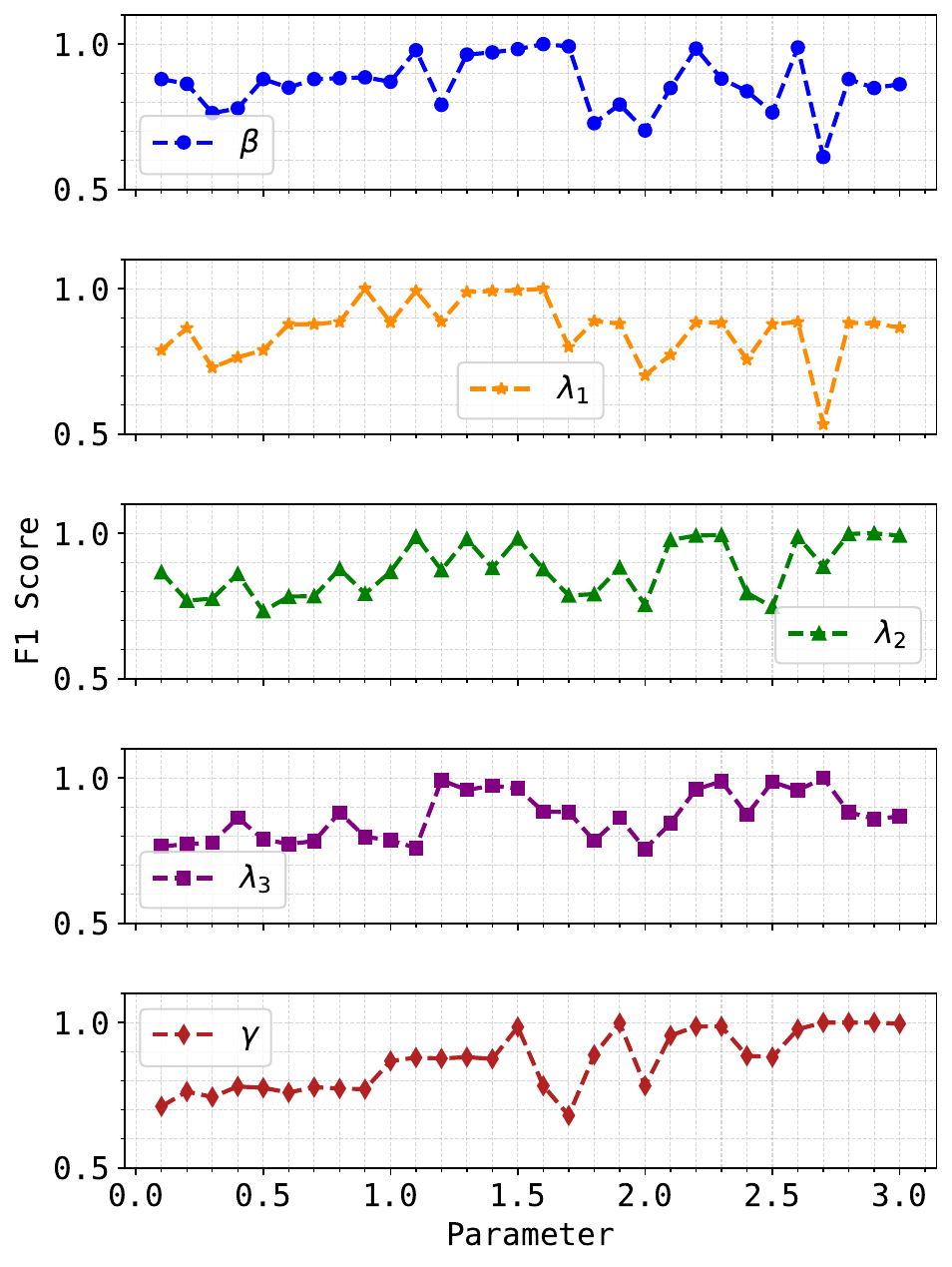}
        \caption{Sensitivity of F1-Score for given parameters for \emph{Ambient Temperature (AT)} dataset.}
        \label{fig:ambient_sa}
    \end{minipage}%
    \hfill
    \begin{minipage}[t]{0.48\textwidth}
        \centering
        \includegraphics[width=\linewidth]{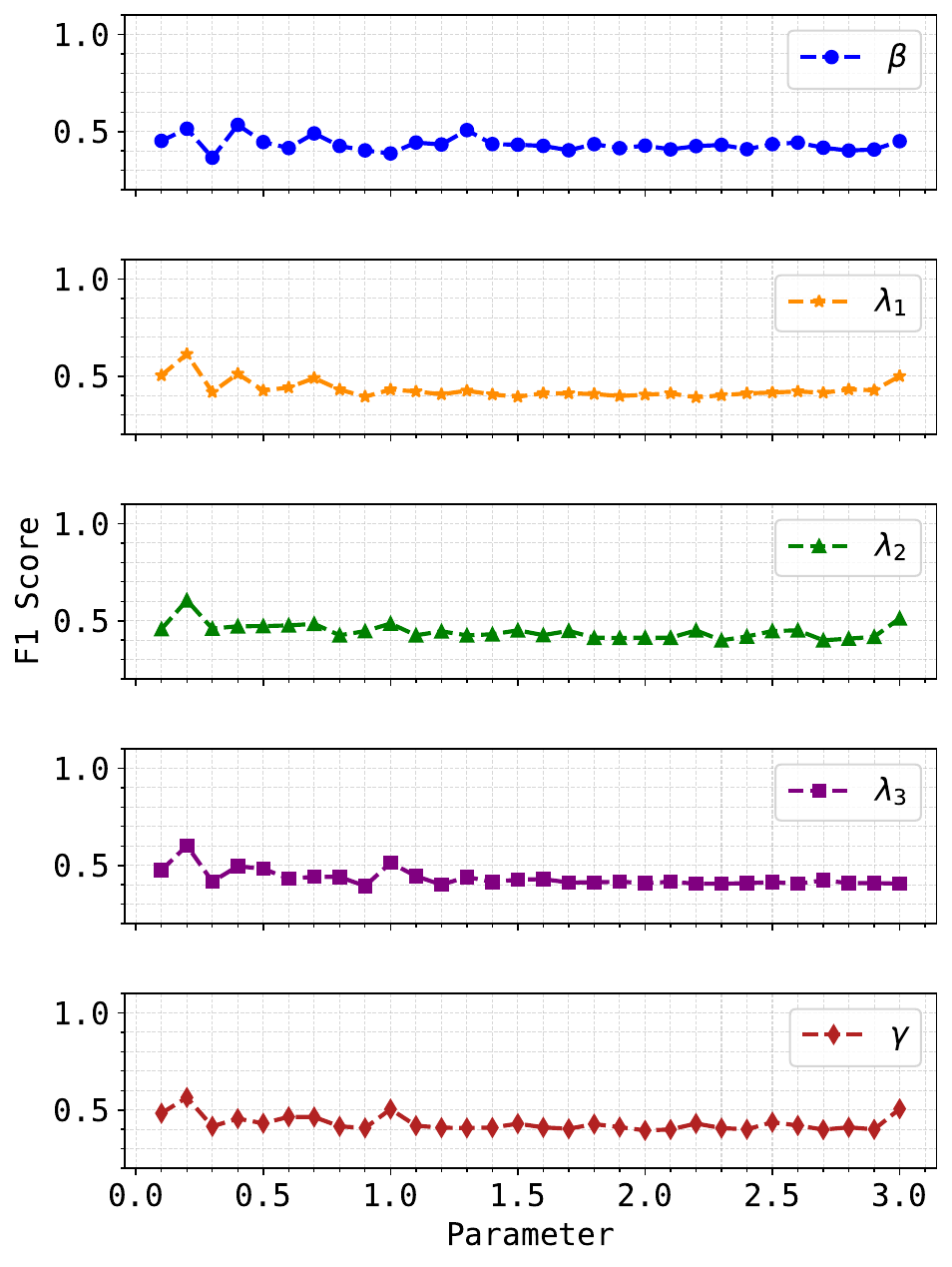}
        \caption{Sensitivity of F1-Score for given parameters for \emph{Machine Temperature (MT)} dataset.}
        \label{fig:machine_sa}
    \end{minipage}
\end{figure}

An interesting observation can be made for MT dataset, as shown in Figure \ref{fig:machine_data}.
There is a very prominent anomaly around the index $4000$ which can be easily detected by a human observer. Although there is no anomaly between index $2000-3000$, \fname{} still highlights it as an anomaly (refer to Anomaly Score in Fig \ref{fig:machine_data}) as it might be a good idea for an observer to pay attention to this dip in machine temperature which can potentially signal a fault. This is also the reason for the F1 score to be lower for this dataset compared to other datasets, as there are some peaks that can potentially be alarming and contribute to false positives. This is also reflected in the sensitivity analysis plot in Figure \ref{fig:machine_sa}. The plot shows minimal variation of F1-Score across all weight parameters. This indicates that the model performance is relatively robust to changes in weights, suggesting that the anomalies in this data set are more prominent and easily distinguishable.

Sensitivity analysis for these datasets thus shows the effectiveness of using a multilevel loss function which integrates reconstruction errors across multiple wavelet levels to capture both coarse and fine grained anomalies.  The variation in sensitivity across different datasets highlights that anomalies manifest differently depending on the underlying signal characteristics. This underlines the necessity of tuning the wavelet-level weights in a dataset-specific manner rather than relying on a fixed configuration for the entire benchmark.  
\section{Related Work}
\label{sec:relatedwork}
Recent advances in deep learning have led to a variety of methods for univariate time series anomaly detection (UTSAD),
particularly in unsupervised settings. These methods can generally be categorized into two groups: \one
forecasting-based approaches that detect anomalies by comparing predicted and actual values and \two reconstruction-based
approaches that learn normal patterns and detect deviations during reconstruction. Table 1 provides a summary of representative deep
learning models across these categories. Details about these methods are presented in the
following sections. 
\begin{table*}[h!]
\centering
\caption{Summary of unsupervised deep learning methods for UTSAD. (AE: Autoencoders, VAE: Variational Autoencoders, GAN: Generative Adversarial Network)}
\renewcommand{\arraystretch}{1.5}
\setlength{\tabcolsep}{2pt} 
\resizebox{0.8\textwidth}{!}{
\begin{tabular}{l l l l l l l}

\toprule
\makecell{\textbf{Methods}} & \textbf{Year} & \textbf{Method Type} & \makecell{\textbf{Architectural} \\ \textbf{Backbone}} & 
 \makecell{\textbf{Architectural } \\ \textbf{Features}} & \makecell{\textbf{Frequency} \\ \textbf{Features}} & \makecell{\textbf{Edge} \\ \textbf{Inference}} \\
\midrule
LSTM-AD (\cite{lstm-ad}) & 2015 & Forecasting & RNN, LSTM &  \XSolidBrush & \XSolidBrush & \XSolidBrush \\

LSTM-RNN (\cite{lstm-ad2}) & 2016 & Forecasting & RNN, LSTM  & \XSolidBrush & \XSolidBrush & \XSolidBrush \\

DeepANT (\cite{deep-ant}) & 2018 & Forecasting & CNN & \XSolidBrush & \XSolidBrush & \XSolidBrush \\
SR-CNN (\cite{srcnn}) & 2019 & Forecasting & CNN & \XSolidBrush & \CheckmarkBold (DFT) & \XSolidBrush \\
\midrule
VAE (\cite{vae}) & 2015 & Reconstruction & VAE & \XSolidBrush & \XSolidBrush & \XSolidBrush \\

EncDec-AD (\cite{lstm-encdec}) & 2016 & Reconstruction & AE & LSTM & \XSolidBrush & \XSolidBrush \\

Donut (\cite{donut}) & 2018 & Reconstruction & VAE & \XSolidBrush & \XSolidBrush & \XSolidBrush
\\
CVAE (\cite{cvae}) & 2020 & Reconstruction & VAE & CNN & \XSolidBrush &  \XSolidBrush
\\

\textbf{VAE-LSTM (\cite{tsvae})} & 2020 & Reconstruction & \textbf{VAE} & LSTM & \XSolidBrush & \XSolidBrush
\\

TAnoGAN (\cite{tanogan}) & 2020  & Reconstruction & GAN & LSTM & \XSolidBrush & \XSolidBrush
\\

\textbf{TADGAN (\cite{tadgan})} & 2020 & Reconstruction & \textbf{GAN} & LSTM, RNN & \XSolidBrush & \XSolidBrush
\\
\textbf{FCVAE (\cite{fcvae})} & 2024 & Reconstruction & \textbf{VAE}  &  \XSolidBrush &  \CheckmarkBold (DFT) & \XSolidBrush
\\
\textbf{\fname{} (Ours)} & 2025 & Reconstruction & \textbf{AE} & CNN & \CheckmarkBold (DWT) & \CheckmarkBold \\
\bottomrule
\end{tabular}
}
\label{tab:rw_summary}
\end{table*}

\subsection{Forecasting-based Methods}
The initial wave of deep learning approaches for UTSAD primarily focused on forecasting-based methods. These
models learned to predict future values from historical observations, with anomalies identified as points where
prediction errors exceeded a threshold. These were based on {\em RNNs}, {\em LSTMs}, {\em CNNs/TCNs} and {\em Transformers}.
LSTM-AD (\cite{lstm-ad}) utilized hierarchical recurrent layers with LSTM units to model temporal
dependencies without requiring labeled data. This approach proved effective for unsupervised anomaly detection,
especially in capturing long-range patterns. Building on this foundation, subsequent work such as (\cite{lstm-ad2})
introduced a simplified LSTM-RNN model that reduced architectural complexity while maintaining reliable performance,
particularly for a sequence of anomalies. As the field evolved, convolutional architectures gained traction due to their
computational efficiency and suitability for parallelization. DeepANT (\cite{deep-ant}) proposed a fully CNN-based
architecture that eliminated recurrent layers, instead emphasizing on local temporal features.  SR-CNN (\cite{srcnn})
incorporated frequency domain information via DFT before applying convolutional layers, aiming to enhance pattern
recognition across time and frequency. 
\\
However, real-world time series data, such as IoT sensor readings, stock prices, or weather data, often exhibit sudden
shifts or unknown influences that can disrupt predictable patterns. As a result, these models tend to generate
increasing prediction errors over time, reducing their reliability for long-term anomaly detection (\cite{anomaly-review}). These limitations motivated a shift toward reconstruction-based approaches, which rely on
modeling normal behavior and identifying deviations in the reconstructed signal, independent of future prediction.
Hence, our focus will be on the more modern reconstruction-based approaches. 

\subsection{Reconstruction-based Methods} 
These methods learn to reconstruct normal time series patterns and detect anomalies based on
deviations from expected behavior. Models based on {\em Autoencoders} (AE), {\em Variational Autoencoders} (VAEs) and
{\em Generative Adversarial Networks} (GANs) have been used for reconstruction-based anomaly detection. EncDec-AD (\cite{lstm-encdec}) demonstrated that training on normal time series alone could enable reliable anomaly detection via
reconstruction error. This approach paved the way for more sophisticated architectures that combined probabilistic
modeling with deep learning. 

VAE-based methods (\cite{vae,cvae,tsvae}) model latent uncertainty and variability using stochastic sampling, but this introduces additional computational overhead due to the need for sampling and KL divergence regularization during both training and inference.  Hybrid models such as
VAE-LSTM (\cite{tsvae}) integrated VAEs with LSTM layers, extracting local representations through the VAE and modeling
long-term dependencies with LSTMs. 
Simultaneously, Generative Adversarial Networks (GANs) gained traction for
generating realistic reconstructions. TanoGAN (\cite{tanogan}) extended the GAN framework to time series by incorporating
LSTM-based generators and demonstrated improved anomaly detection via adversarial training. Similarly,
TadGAN (\cite{tadgan}) employed LSTM-based architectures for both generator and critic networks, effectively capturing
temporal dynamics in an unsupervised setting. GAN-based models are often difficult to train stably and demand higher compute due to repeated forward-backward passes through both networks, making them poorly suited to energy-limited edge settings (\cite{gans}). Moreover, as already highlighted in Section \ref{sec:intro}, subtle anomalies still remain undetected by these reconstruction-based approaches. For instance, LSTMs being used along  with VAE and GAN architectures, are effective at learning sequential patterns, but they often generalize over fine-grained variations due to their reliance on gating mechanisms (e.g., sigmoid and tanh functions), which compress input values into narrow activation ranges. As a result, minor deviations in the input may not sufficiently influence the hidden states, causing subtle anomalies to be overlooked. 

More recently, Transformers (\cite{transformer}) have been explored for
time series anomaly detection, offering improved modeling of long-range dependencies but at the cost of increased
computational complexity. FCVAE (\cite{fcvae}) enhances VAEs for unsupervised anomaly detection by integrating global
and local frequency features to capture periodic patterns and fine-grained trends in univariate time series. Its {\em
target attention} mechanism selects key frequency components, improving short-periodic trend reconstruction. 

In our work, we select baseline methods which represent different styles of
reconstruction based methods. We include VAE-LSTM (parameter count of $700K$), FCVAE (parameter count of $1400K$) and TadGAN (parameter count of $250K$) - methods with different modeling styles, first two are based on
probabilistic latent spaces and the third on adversarial learning. By including these diverse methods, we aim to provide a comprehensive evaluation of our proposed approach against different reconstruction strategies. 

In contrast, \fname{} (parameter count of $97K$) focuses on autoencoder-based reconstruction due to its balance between effectiveness and efficiency.
Unlike VAEs, GANs, or Transformers which introduce additional complexity through stochastic sampling, adversarial
optimization, or attention mechanisms, standard autoencoders offer a lightweight alternative with a compact latent space
and fewer parameters.

\section{Conclusion and Future Work}
\label{sec:conclusion}
We introduced \fname{} (Lightweight Wavelet AutoEncoder), a resource-efficient framework for univariate time-series anomaly detection. \fname{} effectively captures anomalies across different temporal scales without relying on deep or complex architectures by integrating multi-resolution representations from Discrete Wavelet Transform into a compact autoencoder. Our approach not only achieves competitive accuracy on benchmark datasets, but also meets the practical demands of real-world deployment, offering low latency and minimal power consumption on Jetson Nano, highlighting its suitability for real-time, energy-efficient anomaly detection in resource-constrained environments.

For future work, we aim to study the use of different wavelet families and observe how the model complexity can grow due to change in the input sequence padding and edge effects introduced by other wavelets. We also aim to explore adaptive weighting strategies for weight parameters ($\beta,\lambda,\gamma$) of our loss function.

\bibliographystyle{unsrt} 
\bibliography{references}

@article{b1,
  title={A theory for multiresolution signal decomposition: the wavelet representation},
  author={Mallat, Stephane G},
  journal={IEEE transactions on pattern analysis and machine intelligence},
  volume={11},
  number={7},
  pages={674--693},
  year={2002},
  publisher={Ieee}
}

@inproceedings{b2,
  title={Evaluating real-time anomaly detection algorithms--the Numenta anomaly benchmark},
  author={Lavin, Alexander and Ahmad, Subutai},
  booktitle={2015 IEEE 14th international conference on machine learning and applications (ICMLA)},
  pages={38--44},
  year={2015},
  organization={IEEE}
}

@article{edge,
  title={Smart factory of industry 4.0: Key technologies, application case, and challenges},
  author={Chen, Baotong and Wan, Jiafu and Shu, Lei and Li, Peng and Mukherjee, Mithun and Yin, Boxing},
  journal={Ieee Access},
  volume={6},
  pages={6505--6519},
  year={2017},
  publisher={IEEE}
}

@inproceedings{tsvae,
  title={Anomaly detection for time series using vae-lstm hybrid model},
  author={Lin, Shuyu and Clark, Ronald and Birke, Robert and Sch{\"o}nborn, Sandro and Trigoni, Niki and Roberts, Stephen},
  booktitle={ICASSP 2020-2020 IEEE International Conference on Acoustics, Speech and Signal Processing (ICASSP)},
  pages={4322--4326},
  year={2020},
  organization={Ieee}
}

@article{wave-rora,
  title={WaveRoRA: Wavelet Rotary Route Attention for Multivariate Time Series Forecasting},
  author={Liang, Aobo and Sun, Yan and Guizani, Nadra},
  journal={arXiv preprint arXiv:2410.22649},
  year={2024}
}

@article{dwt-psi,
  title={Discrete wavelet transform-based time series analysis and mining},
  author={Chaovalit, Pimwadee and Gangopadhyay, Aryya and Karabatis, George and Chen, Zhiyuan},
  journal={ACM Computing Surveys (CSUR)},
  volume={43},
  number={2},
  pages={1--37},
  year={2011},
  publisher={ACM New York, NY, USA}
}

@article{dwt-eq,
  title={Mathematical representations of 1D, 2D and 3D wavelet transform for image coding},
  author={Ravichandran, D and Nimmatoori, Ramesh and Ahamad, M Gulam},
  journal={Int. J. Adv. Comput. Theory Eng},
  volume={5},
  number={3},
  pages={20--27},
  year={2016}
}

@inproceedings{haar,
  title={The Haar wavelet transform in the time series similarity paradigm},
  author={Struzik, Zbigniew R and Siebes, Arno},
  booktitle={European Conference on Principles of Data Mining and Knowledge Discovery},
  pages={12--22},
  year={1999},
  organization={Springer}
}

@article{daubechies,
  title={Daubechies wavelets and mathematica},
  author={Rowe, Alistair CH and Abbott, Paul C},
  journal={Computers in Physics},
  volume={9},
  number={6},
  pages={635--648},
  year={1995},
  publisher={American Institute of Physics}
}

@article{anomaly-review,
  title={Deep learning for time series anomaly detection: A survey},
  author={Zamanzadeh Darban, Zahra and Webb, Geoffrey I and Pan, Shirui and Aggarwal, Charu and Salehi, Mahsa},
  journal={ACM Computing Surveys},
  volume={57},
  number={1},
  pages={1--42},
  year={2024},
  publisher={ACM New York, NY}
}

@article{wavelet-theory,
  title={Wavelet theory and applications: a literature study},
  author={Merry, RJE},
  year={2005},
  publisher={Technische Universiteit Eindhoven}
}

@inproceedings{eval,
  title={Time-series anomaly detection service at microsoft},
  author={Ren, Hansheng and Xu, Bixiong and Wang, Yujing and Yi, Chao and Huang, Congrui and Kou, Xiaoyu and Xing, Tony and Yang, Mao and Tong, Jie and Zhang, Qi},
  booktitle={Proceedings of the 25th ACM SIGKDD international conference on knowledge discovery \& data mining},
  pages={3009--3017},
  year={2019}
}

@inproceedings{tadgan,
  title={Tadgan: Time series anomaly detection using generative adversarial networks},
  author={Geiger, Alexander and Liu, Dongyu and Alnegheimish, Sarah and Cuesta-Infante, Alfredo and Veeramachaneni, Kalyan},
  booktitle={2020 ieee international conference on big data (big data)},
  pages={33--43},
  year={2020},
  organization={IEEE}
}

@inproceedings{fcvae,
  title={Revisiting vae for unsupervised time series anomaly detection: A frequency perspective},
  author={Wang, Zexin and Pei, Changhua and Ma, Minghua and Wang, Xin and Li, Zhihan and Pei, Dan and Rajmohan, Saravan and Zhang, Dongmei and Lin, Qingwei and Zhang, Haiming and others},
  booktitle={Proceedings of the ACM Web Conference 2024},
  pages={3096--3105},
  year={2024}
}

@article{transformer,
  title={Anomaly transformer: Time series anomaly detection with association discrepancy},
  author={Xu, Jiehui and Wu, Haixu and Wang, Jianmin and Long, Mingsheng},
  journal={arXiv preprint arXiv:2110.02642},
  year={2021}
}

@inproceedings{lstm-ad,
  title={Long short term memory networks for anomaly detection in time series},
  author={Malhotra, Pankaj and Vig, Lovekesh and Shroff, Gautam and Agarwal, Puneet and others},
  booktitle={Proceedings},
  volume={89},
  number={9},
  pages={94},
  year={2015}
}

@inproceedings{lstm-ad2,
  title={Collective anomaly detection based on long short-term memory recurrent neural networks},
  author={Bontemps, Lo{\"\i}c and Cao, Van Loi and McDermott, James and Le-Khac, Nhien-An},
  booktitle={Future Data and Security Engineering: Third International Conference, FDSE 2016, Can Tho City, Vietnam, November 23-25, 2016, Proceedings 3},
  pages={141--152},
  year={2016},
  organization={Springer}
}

@inproceedings{srcnn,
  title={Time-series anomaly detection service at microsoft},
  author={Ren, Hansheng and Xu, Bixiong and Wang, Yujing and Yi, Chao and Huang, Congrui and Kou, Xiaoyu and Xing, Tony and Yang, Mao and Tong, Jie and Zhang, Qi},
  booktitle={Proceedings of the 25th ACM SIGKDD international conference on knowledge discovery \& data mining},
  pages={3009--3017},
  year={2019}
}

@article{deep-ant,
  title={DeepAnT: A deep learning approach for unsupervised anomaly detection in time series},
  author={Munir, Mohsin and Siddiqui, Shoaib Ahmed and Dengel, Andreas and Ahmed, Sheraz},
  journal={Ieee Access},
  volume={7},
  pages={1991--2005},
  year={2018},
  publisher={IEEE}
}

@article{lstm-encdec,
  title={LSTM-based encoder-decoder for multi-sensor anomaly detection},
  author={Malhotra, Pankaj and Ramakrishnan, Anusha and Anand, Gaurangi and Vig, Lovekesh and Agarwal, Puneet and Shroff, Gautam},
  journal={arXiv preprint arXiv:1607.00148},
  year={2016}
}

@article{vae,
  title={Variational autoencoder based anomaly detection using reconstruction probability},
  author={An, Jinwon and Cho, Sungzoon},
  journal={Special lecture on IE},
  volume={2},
  number={1},
  pages={1--18},
  year={2015}
}

@inproceedings{donut,
  title={Unsupervised anomaly detection via variational auto-encoder for seasonal kpis in web applications},
  author={Xu, Haowen and Chen, Wenxiao and Zhao, Nengwen and Li, Zeyan and Bu, Jiahao and Li, Zhihan and Liu, Ying and Zhao, Youjian and Pei, Dan and Feng, Yang and others},
  booktitle={Proceedings of the 2018 world wide web conference},
  pages={187--196},
  year={2018}
}

@inproceedings{cvae,
  title={Unsupervised Anomaly Detection in High-Dimensional Flight Data Using Convolutional Variational Auto-Encoder},
  author={Memarzadeh, Milad and Matthews, Bryan and Avrekh, Ilya and Weckler, Daniel},
  booktitle={26th SIGKDD Conference on Knowledge Discovery and Data Mining},
  number={ARC-E-DAA-TN77606},
  year={2020}
}

@inproceedings{tanogan,
  title={TAnoGAN: Time series anomaly detection with generative adversarial networks},
  author={Bashar, Md Abul and Nayak, Richi},
  booktitle={2020 IEEE Symposium Series on Computational Intelligence (SSCI)},
  pages={1778--1785},
  year={2020},
  organization={IEEE}
}

@inproceedings{deep-edge-bench,
  title={DeepEdgeBench: Benchmarking deep neural networks on edge devices},
  author={Baller, Stephan Patrick and Jindal, Anshul and Chadha, Mohak and Gerndt, Michael},
  booktitle={2021 IEEE International Conference on Cloud Engineering (IC2E)},
  pages={20--30},
  year={2021},
  organization={IEEE}
}

@article{traditional-methods,
  title={Anomaly detection: A survey},
  author={Chandola, Varun and Banerjee, Arindam and Kumar, Vipin},
  journal={ACM computing surveys (CSUR)},
  volume={41},
  number={3},
  pages={1--58},
  year={2009},
  publisher={ACM New York, NY, USA}
}

@article{knn,
  title={On the time series $ k $-nearest neighbor classification of abnormal brain activity},
  author={Chaovalitwongse, Wanpracha Art and Fan, Ya-Ju and Sachdeo, Rajesh C},
  journal={IEEE Transactions on Systems, Man, and Cybernetics-Part A: Systems and Humans},
  volume={37},
  number={6},
  pages={1005--1016},
  year={2007},
  publisher={IEEE}
}

@inproceedings{spot,
  title={Anomaly detection in streams with extreme value theory},
  author={Siffer, Alban and Fouque, Pierre-Alain and Termier, Alexandre and Largouet, Christine},
  booktitle={Proceedings of the 23rd ACM SIGKDD international conference on knowledge discovery and data mining},
  pages={1067--1075},
  year={2017}
}

@article{gans,
  title={Generative adversarial networks (GANs): An overview of theoretical model, evaluation metrics, and recent developments},
  author={Salehi, Pegah and Chalechale, Abdolah and Taghizadeh, Maryam},
  journal={arXiv preprint arXiv:2005.13178},
  year={2020}
}

@misc{jetson,
  author       = {{NVIDIA}},
  title        = {Jetson Nano Developer Kit},
  year         = {2021},
  howpublished = {\url{https://developer.nvidia.com/embedded/jetson-nano-developer-kit}},
  note         = {Accessed: 2025-07-05}
}

@misc{yahoo,
  author       = {{Yahoo}},
  title        = {Yahoo Anomaly Detection Dataset},
  howpublished = {\url{https://webscope.sandbox.yahoo.com/catalog.php?datatype=s}},
  note         = {Accessed: 2025-07-05},
  year         = {n.d.}
}
\end{document}